\documentclass[letterpaper]{article}
\usepackage{aaai2026}  % DO NOT CHANGE THIS

\usepackage{times}  % DO NOT CHANGE THIS
\usepackage{helvet}  % DO NOT CHANGE THIS
\usepackage{courier}  % DO NOT CHANGE THIS
\usepackage[hyphens]{url}  % DO NOT CHANGE THIS
\usepackage{graphicx} % DO NOT CHANGE THIS
\usepackage{natbib}  % DO NOT CHANGE THIS AND DO NOT ADD ANY OPTIONS TO IT
\usepackage{caption} % DO NOT CHANGE THIS AND DO NOT ADD ANY OPTIONS TO IT
\usepackage{algorithm}
\usepackage{algorithmic}
\usepackage{newfloat}
\usepackage{listings}
\usepackage{float}
\usepackage{pifont}
\usepackage{tcolorbox}
\usepackage{multirow}
\usepackage[table]{xcolor}
\usepackage{graphicx}
\usepackage{booktabs}
\usepackage{nameref}
\DeclareCaptionStyle{ruled}{labelfont=normalfont,labelsep=colon,strut=off} % DO NOT CHANGE THIS
\floatstyle{ruled}
\newfloat{listing}{tb}{lst}{}
\floatname{listing}{Listing}
\definecolor{olivegreen}{RGB}{30,130,47}
\definecolor{custombrown}{RGB}{200,30,0}

\title{The Failure Happens Before the Drift: The Social Dynamics of Values in LLM Agent Societies}

\author{
    Farah Atif\textsuperscript{\rm 1},
    Sougata Saha\textsuperscript{\rm 1},
    Monojit Choudhury\textsuperscript{\rm 1},
}
\affiliations{
    \textsuperscript{\rm 1}Mohamed Bin Zayed University of Artificial Intelligence\\
    \{farah.atif, sougata.saha, monojit.choudhury\}@mbzuai.ac.ae
}

\usepackage{bibentry}
\begin{document}

\maketitle

% ---------------------------------------------------------------
%  ABSTRACT
% ---------------------------------------------------------------
\begin{abstract}

Large Language Model (LLM)-based agents are increasingly used as proxies for human participants in social science research, yet it remains unclear whether they can faithfully simulate diverse and conflicting human value systems. We present a World Values Survey (WVS)-grounded simulation framework where culturally diverse agents with different communication styles engage in longitudinal, value-laden discussions. Across approximately 4,000 conversations involving 1,200 personas, 15 topics, and three models (GPT-4o, Gemini-2.5-Flash, and Gemma-4-E4B), we evaluate value faithfulness, value drift, and conversational realism. We find that more than 50\% of personas fail to express their assigned WVS profiles from the outset, while 2-7\% drift after repeated conversations. Ablations removing demographic details improve faithfulness for some models but do not change the broader trend: simulated value distributions still systematically deviate from the assigned WVS profiles. Compared to human discussions, simulated dialogues show a different trade-off between stylistic consistency and semantic diversity, often producing content-wise varied but stylistically repetitive exchanges. These findings suggest that current LLM agents can generate plausible conversations, but remain limited proxies for representing and preserving diverse human value profiles over time.

% An ablation removing persona descriptions improves value faithfulness from 44% to 71% in larger models, suggesting interference between maintaining a value system and performing a cultural identity. Our study indicates that current LLMs are not yet reliable proxies for social science research because they struggle to consistently instantiate and sustain diverse human value profiles.
% \textcolor{red}{Persona-conditioned discussion of policy topics systematically pulls simulated populations toward Survival framings, a directional bias that runs counter to the WEIRD defaults documented in single-turn elicitation, and that suggests a distinct failure mode in multi-turn discourse.}
   
\end{abstract}

% ---------------------------------------------------------------
%  1. INTRODUCTION
% ---------------------------------------------------------------
\section{Introduction}
% There is a growing interest in using large language models (LLMs) as stand-ins for human populations. From simulating individuals that mirror specific demographics or personality traits \citep{argyle2023out,tjuatja2024llms,wang2025evaluating} to modeling collective phenomena such as group dynamics and organizational behavior \citep{park2024generative,zhou2024sotopia}, LLMs are increasingly positioned as scalable proxies for human judgment, opinion, and behavior \citep{choi2026overstating,tjuatja2024llms,park2024generative}. However, real-world societies are defined not only by demographic diversity but also by diverse and often conflicting value systems. The validity of these simulations, therefore, depends on whether LLM agents can reliably instantiate and sustain culturally grounded value profiles during interaction.

There is a growing interest in using LLMs as stand-ins for human populations. Prior work has used LLMs to simulate individuals with specific demographic attributes or personality traits \citep{argyle2023out,tjuatja2024llms,wang2025evaluating}, as well as collective phenomena such as group dynamics and organizational behavior \citep{park2024generative,zhou2024sotopia}. In these settings, LLMs are increasingly positioned as scalable proxies for human judgment, opinion, and behavior \citep{choi2026overstating,tjuatja2024llms,park2024generative}. However, real-world societies are shaped not only by demographic diversity but also by diverse and often conflicting value systems\cite{inbook}. Therefore, the validity of such simulations depends on whether LLM agents can reliably instantiate and sustain culturally grounded value profiles during interaction.

% Most existing evaluations of LLM value alignment are static, where models are typically prompted with survey questions, Likert-scale inventories, or value benchmarks, and their responses are compared against human distributions \citep{santurkar2023whose,johnson2022ghost,dwivedi2023eticor}. Such evaluations have shown LLMs to often gravitate toward Western, educated, liberal orientations regardless of persona conditioning. Yet these evaluations remain fundamentally static, as they measure what a model states in isolation, not whether assigned value systems are successfully instantiated in conversational behavior, nor whether they remain stable through repeated social interaction and contestation.

Most existing evaluations of LLM value alignment are static: models answer survey items, Likert-scale inventories, or value benchmarks, and their responses are compared against human distributions \citep{santurkar2023whose,johnson2022ghost,dwivedi-etal-2023-eticor}. These studies show that LLMs often gravitate toward Western, educated, and liberal orientations, even under persona conditioning \citep{dwivedi-etal-2023-eticor,johnson2022ghost}. However, such pointwise evaluations measure values stated in isolation, as they do not test whether assigned values are expressed in conversation or sustained through repeated interaction, disagreement, and contestation.

% Most existing evaluations of LLM value alignment are static. Models are typically prompted with survey questions, Likert-scale inventories, or value benchmarks, and their responses are compared against human distributions \citep{santurkar2023whose,johnson2022ghost,dwivedi-etal-2023-eticor}. These studies show that LLMs often gravitate toward Western, educated, and liberal orientations, even under persona conditioning \cite{dwivedi-etal-2023-eticor,johnson2022ghost}. However, such ``pointwise'' evaluations measure what a model states in isolation. Neither do they test whether assigned value systems are expressed in conversational behavior, nor whether these values remain stable through repeated social interaction, disagreement, and contestation.

\begin{figure}[t]
    \centering
    \includegraphics[width=.46\textwidth, height=6cm]{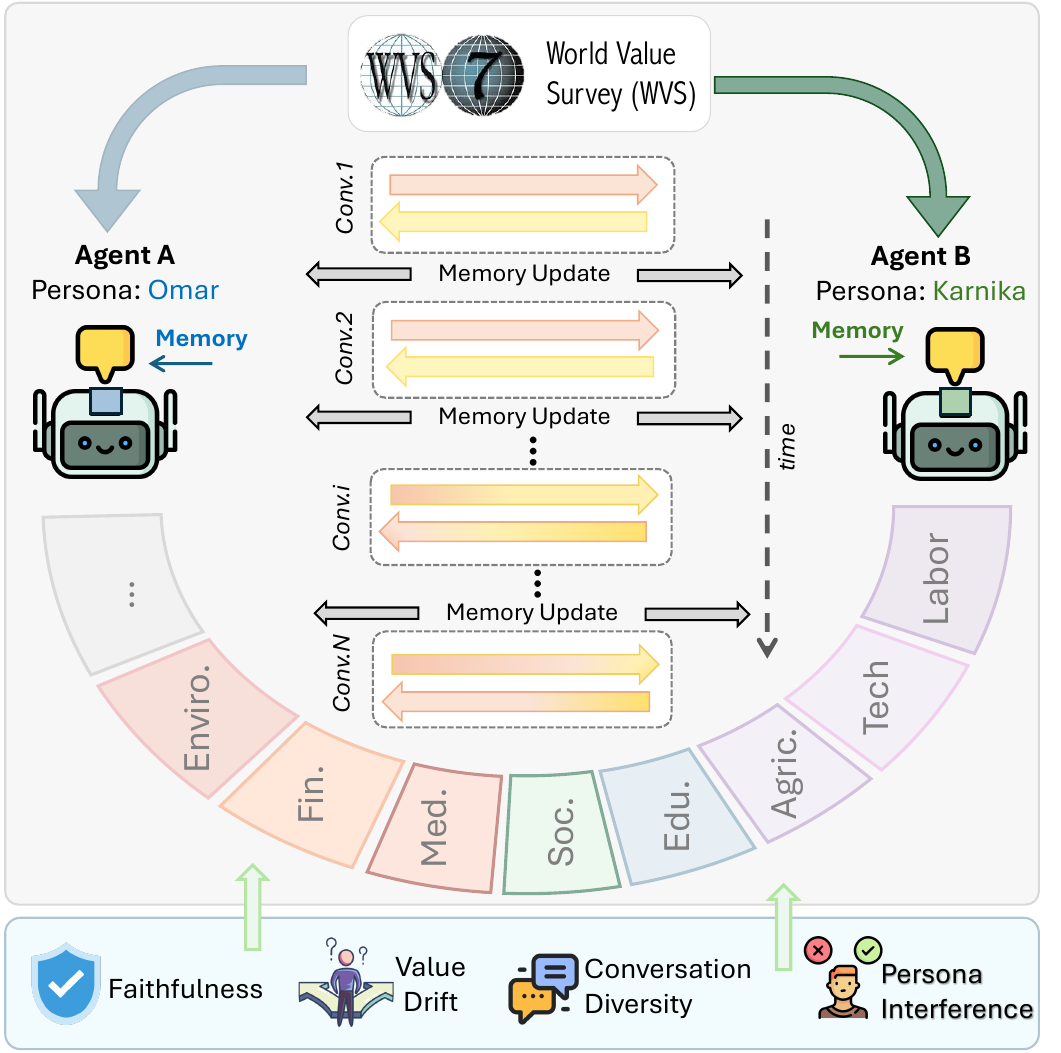}
    \caption{Overview of the proposed WVS-grounded multi-agent simulation framework, where culturally conditioned LLM personas engage in repeated value-driven discussions across diverse topics and are evaluated for value faithfulness, value drift, and conversational realism.}
    \vspace{-0.5cm}
    % \caption{Overview of the proposed WVS-grounded multi-agent simulation framework, where culturally conditioned LLM personas engage in repeated value-driven discussions across diverse social domains. The framework tracks memory accumulation and evaluates simulations through four dimensions: value faithfulness, value drift, conversational diversity, and persona interference.}
    \label{fig:intro}
\end{figure}

% In this work, we distinguish between two fundamental properties of value-driven LLM societies: \textbf{Instantiation} and \textbf{Retention}. \textit{Instantiation} measures whether an agent initially expresses the profile of cultural value assigned when first deployed. \textit{Retention} measures whether, once instantiated, that value orientation remains recoverable across repeated multi-turn discussions. Although existing benchmarks can partially evaluate instantiation through static responses, they cannot capture retention under sustained interaction. More importantly, they often implicitly assume that initialization is successful in the first place.

In this work, we distinguish between two fundamental properties of value-driven LLM societies: \textbf{Instantiation} and \textbf{Retention}. \textit{Instantiation} measures whether an agent expresses its assigned cultural value profile when first deployed. \textit{Retention} measures whether, once instantiated, that value orientation remains recoverable across repeated multi-turn discussions. Existing benchmarks mostly evaluate instantiation through static responses, but do not capture retention under sustained interaction. %More importantly, they often assume that persona initialization succeeds in the first place.

% To study these questions, we introduce a large-scale simulation framework grounded in the World Values Survey (WVS) \citep{haerpfer2022world}. Personas are sampled from WVS Round Seven across eight cultural regions and assigned positions on the two Inglehart-Welzel cultural dimensions \textit{Traditional vs. Secular-Rational} and \textit{Survival vs. Self-Expression}. Agents engage in repeated multi-turn discussions on 15 value-laden topics spanning labor, technology, agriculture, and social policy. After each interaction, agents update a structured memory summarizing the discussion and its inferred effect on their value orientation, allowing conversational experiences to accumulate over time. We evaluate simulations across GPT-4o, Gemini-2.5-Flash, and Gemma-4-E4B along four dimensions: value faithfulness, value drift, conversational diversity, and persona interference.

To study these questions, we introduce a large-scale simulation framework\footnote{Code: \url{https://github.com/mbzuai-nlp/wvs-value-simulation}} \footnote{Data: \url{https://huggingface.co/datasets/MBZUAI/WVS-Value-Simulation}} grounded in the WVS \citep{haerpfer2022world} (Figure~\ref{fig:intro}). We sample personas from WVS Wave 7 across eight cultural regions and assign them positions along the two Inglehart-Welzel cultural dimensions \citep{Inglehart_Welzel_2005}: \textit{Traditional--Secular-Rational} and \textit{Survival--Self-Expression}. Agents then engage in repeated multi-turn discussions on 15 value-laden topics spanning labor, technology, agriculture, education, and social policy. After each interaction, agents update a structured memory that summarizes the discussion and its inferred effect on their value orientation, allowing conversational experiences to accumulate over time. We evaluate simulations across GPT-4o, Gemini-2.5-Flash, and Gemma-4-E4B for value faithfulness, value drift, and conversational quality.

% Our central finding is that the primary failure in LLM social simulation occurs before conversational drift begins. Across all evaluated models, fewer than 7\% of personas exhibit substantial value drift over their full conversational histories. However, a much larger proportion fail to correctly instantiate their assigned WVS value profiles from the outset. We further find that demographic persona conditioning often reduces value faithfulness, and that persuasive communication style is strongly correlated with unfaithfulness. Additionally, value trajectories reveal systematic distortions in the simulated value space: moderate value profiles tend to collapse toward Traditional/Survival orientations, while several rare value combinations present in the original WVS distribution become disproportionately amplified after repeated interactions. Finally, our conversational diversity analysis shows that models differ systematically in their balance between style and content variation: GPT-4o produces the most stylistically consistent outputs but also shows high content similarity within topics, while Gemma-4-E4B exhibits the greatest stylistic and content diversity. Gemini-2.5-Flash generally falls between these two patterns, often overlapping with the human baseline in style while showing relatively varied content distributions.

Our analysis suggests that the primary failure in all LLM social simulations often occurs within the first two conversations: in nearly 50\% of cases, LLM agents fail to exhibit their assigned WVS value profiles from the outset. Ablation studies further show that removing demographic details from the persona increases value faithfulness, while persuasive communication style is strongly correlated with unfaithfulness. Across all evaluated models, fewer than 7\% of personas exhibit substantial value drift over their full conversational histories. However, the final value systems reflected in these simulations systematically deviate from real-world distributions, where moderate value profiles tend to collapse toward Traditional/Survival orientations, while several rare value combinations in the original WVS distribution become disproportionately amplified after repeated interactions.

Finally, our conversational analysis shows that models differ systematically in how they balance style, as reflected by dialogue acts \cite{stolcke-etal-2000-dialogue}, and content, as measured using embedding similarity \cite{DBLP:journals/corr/abs-1301-3781}. GPT-4o produces the most stylistically similar outputs relative to existing human dialogue corpora, but also shows high content similarity within topics. Gemma-4-E4B exhibits the greatest stylistic and content diversity, while Gemini-2.5-Flash generally falls between these two patterns, often overlapping with the human baseline in style while producing relatively varied content distributions.

% This suggests that current concerns about long-term value instability in LLM societies may be secondary to a more fundamental problem of weak initial persona grounding.

These findings have important implications for the use of LLM agents in computational social science. If simulated personas do not reliably instantiate culturally grounded value systems, then downstream results from multi-agent simulations may reflect artifacts of model priors and alignment behavior rather than the intended population distributions. Our work, therefore, reframes the evaluation of LLM societies away from isolated survey responses and toward longitudinal behavioral consistency under interaction.

\section{Simulation Design}
\begin{figure*}[t]
    \centering
    \includegraphics[width=\textwidth]{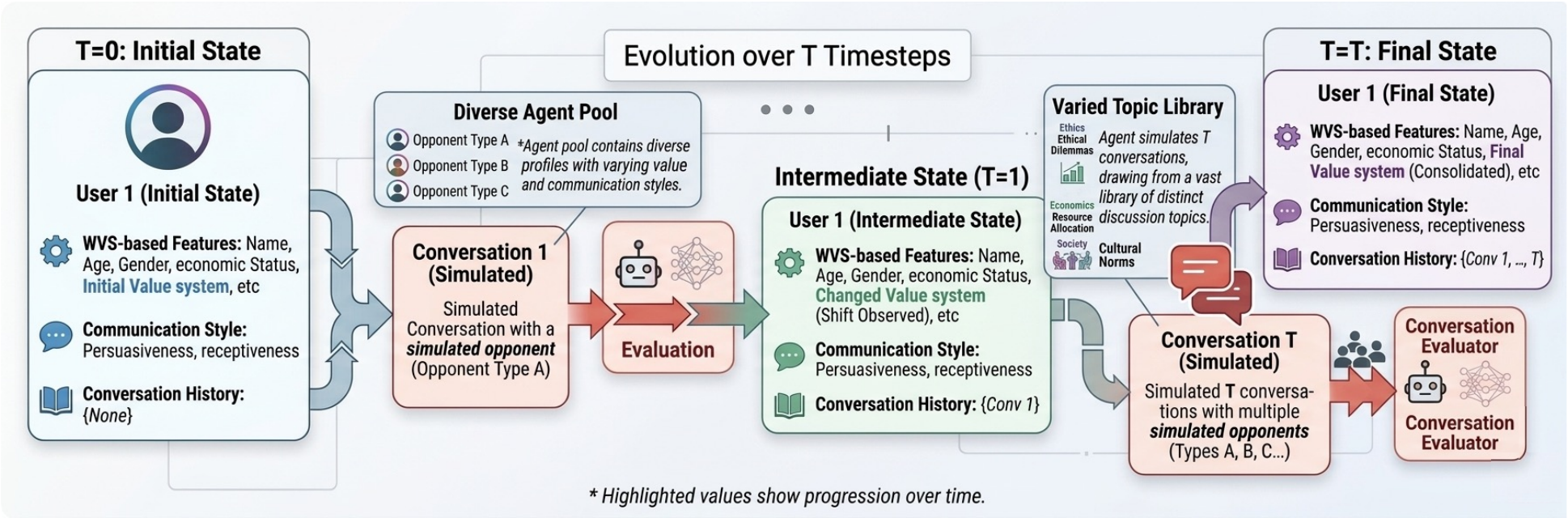}
    \caption{Overview of the WVS-grounded simulation pipeline, where diverse personas engage in repeated value-laden interactions, update their histories and value states, and are analyzed for value evolution and behavioral consistency.}
    \vspace{-0.4cm}
    % \caption{Overview of the WVS-grounded multi-agent simulation pipeline. WVS-grounded personas with diverse demographic and communication profiles engage in repeated multi-turn interactions with heterogeneous opponents across value-laden discussion topics. Over time, agents update their conversational histories and value states, enabling the analysis of value evolution, conversational influence, and longitudinal behavioral consistency.}
    \label{fig:pipeline}
\end{figure*}

Here we describe our end-to-end simulation pipeline in detail. An illustration of the same is presented in Figure \ref{fig:pipeline}.

\subsection{Participant Sampling and Persona Initialization}
Each participant is defined by three broad aspects: their socio-demographic features, value system, and communication style. We derived the participant profiles from the WVS Wave 7 data \cite{WVSWave7}, which spans $\sim$90,000 participants from 66 countries. The WVS is particularly well-suited for this purpose because it is one of the most comprehensive and widely used cross-national surveys of human beliefs, values, and social attitudes \cite{haerpfer2022world, Inglehart_Welzel_2005}. Its broad geographic coverage and standardized survey design make it possible to construct personas that vary not only across countries but also across demographic groups, social positions, and value orientations within countries. Moreover, because WVS Wave 7 contains each participant's responses to $\sim$290 questions pertaining to different aspects of life, such as social values, well-being, religion, economy, political interests, and demographics, it provides a richer basis for persona initialization than datasets that only contain coarse demographic attributes.

For our purposes, we preserved the following \textbf{socio-demographic features} for each individual: age, gender, country of residence, region (urban/rural), known languages, immigration status (immigrant/non-immigrant), household status, education profile of the individual and household, employment profile of the individual and household, economic status, religion, and ethnicity, as detailed in the Persona creation prompt: Part I (in Appendix). These features allow the simulations to condition conversations on concrete social and demographic contexts, rather than relying on abstract or stereotyped representations of cultural groups.

We filtered out participants who did not have values for these features, resulting in $\sim$50,000 participants. We further filtered out participants who belonged to regions (defined by country and urban/rural combination) with $<$50 participants. This filtering was done to ensure sufficient diversity within each region, so that different types of conversations could be simulated without overfitting a region to a small number of observed individuals. From the remaining participants, we randomly sampled $\sim$1,200 final participants for the simulation. The final participant set was limited to 1,200 due to cost-related constraints, since simulating and evaluating longitudinal conversations is costly.

We inferred participants' \textbf{value systems} on a 3-point scale - high, medium, and low - along two bipolar dimensions: ``Traditional--Secular-rational" and ``Survival--Self-expression". These dimensions were inferred based on participants' answers to the 10 indicator questions used to create the Inglehart-Welzel Cultural Map \cite{Inglehart_Welzel_2005}. The map is widely used in comparative social science to summarize cross-national variation in values, while still allowing individual participants within the same country or cultural region to occupy different positions along the two dimensions. Thus, the inferred value systems provide a compact representation of participants' normative and social orientations without reducing them solely to their country of residence. Furthermore, we used GPT-4o to create realistic and culturally appropriate synthetic names for each participant, and created a succinct 100-word bio comprising their demographic features, as detailed in the Persona creation prompt: Part II (in Appendix).

In addition to socio-demographic features and value systems, we equipped each persona with a \emph{communication style}. Communication style is pertinent for our simulations because longitudinal conversations are shaped not only by what participants believe, but also by how they express disagreement, evaluate new information, and respond to competing viewpoints. Prior work on persuasion and conversational receptiveness shows that individuals differ in their willingness to engage with opposing views, their tendency to evaluate arguments impartially, and their use of language to signal openness or resistance during disagreement \cite{minson2022receptiveness, yeomans2020conversational}. Similarly, persuasion research suggests that the effectiveness and form of persuasive interaction depend on individual differences in how people present and process arguments \cite{cacioppo1982need, petty2012communication}. Therefore, modeling communication style allows our simulations to capture variation in conversational dynamics that would not be represented by demographic features or value systems alone.

We define communication style using two axes: \emph{persuasiveness} and \emph{receptiveness}. A high persuasiveness indicates a propensity to convince others and guide them toward one's perspective, whereas a low persuasiveness is reflected by presenting one's arguments more neutrally, without an evident attempt to sway others toward a specific viewpoint. A high receptiveness is reflected by being open to diverse ideas and perspectives, and by thoughtfully considering arguments without demanding immediate evidence or justification. In contrast, low receptiveness is reflected by critically evaluating arguments, requiring substantial evidence and logical consistency before considering alternative viewpoints. Together, persuasiveness and receptiveness define four regimes of communication: both high, both low, high persuasiveness with low receptiveness, and low persuasiveness with high receptiveness.

Importantly, we do not treat communication style as deterministically implied by a participant's cultural background or value system. Although culture can shape norms of communication, cross-cultural research also shows substantial individual-level variation within cultures, and some studies find that individual-level factors are stronger predictors of communication preferences than country-level cultural categories \cite{gudykunst1996influence, park2012individual}. This motivates treating communication style as an additional axis of persona variation rather than deriving it directly from cultural region or WVS-inferred values. Accordingly, each persona was randomly assigned one of the four communication styles. This random assignment helps avoid essentializing cultural groups while allowing the simulations to explore how similar socio-demographic and value profiles may produce different conversational trajectories under different interactional dispositions.

Overall, the $\sim$1,200 participants span eight cultural regions: African-Islamic, Latin America, West \& South Asia, Confucian, English-speaking, Orthodox Europe, Catholic Europe, and Protestant Europe. Their personas were defined by their WVS-based socio-demographic features, inferred value systems, and randomly assigned communication style.

\subsection{Discussion Format and Topic}
Unlike general chitchat, where value-laden exchanges may occur only sparsely, we simulate discussions around a fixed set of occupation-relevant topics designed to elicit disagreement, justification, and perspective-taking. This setup follows prior work on online deliberation and persuasion, where participants respond to explicit claims or prompts rather than engaging in open-ended conversation \cite{tan2016winning, hidey2017analyzing}. It also resembles structured debate platforms such as Kialo\footnote{\url{https://www.kialo.com/}}, where discussions are organized around central claims and supporting or opposing arguments. Such a structure is important for our setting because it increases the likelihood that participants express values, trade-offs, and normative commitments, while still allowing them to respond naturally.

Participants were grouped into 5 occupation-based groups, determined from the WVS dataset: ``Farm Workers, Farm Owners, Semi-Skilled, and Unskilled Workers'', ``Professional, Technical, and Skilled Workers'', ``Service, Sales, and Clerical Roles'', ``Students'', and ``Unemployed, Retired, and Housewife Roles''. Each participant belonged to exactly one group, and the groups were approximately balanced in size. As shown in Table~\ref{tab:topics}, we manually created three occupation-relevant discussion topics per group, yielding 15 topics in total. The topics reflect contemporary issues in global news and public discourse, and were designed to be broadly relevant within each occupational group regardless of country, religion, ethnicity, or value orientation. This occupation-conditioned design supports realistic engagement by grounding discussions in issues tied to participants' work, education, household roles, or livelihoods.

Each topic consists of a main claim and five discussion aspects. The main claim provides a concrete position for participants to respond to, while the aspects constrain the space of possible responses to issues that are substantively relevant to the topic. This design follows findings from research on asynchronous online discussions, which suggests that structured prompts can reduce ambiguity, focus participation, and support more substantive discourse \cite{gao2013designing, hew2012student}. Constraining the discussion in this way also helps mitigate common issues in simulated conversations, such as topical drift, generic responses, and repetitive content.

We framed each interaction as a private discussion between two members of a social-media group related to their occupational background. One member had posted a claim about the topic, and the other participant responded. This framing reflects common forms of online discussion, where people engage in asynchronous, text-based exchanges within interest-, identity-, or occupation-based groups \cite{boyd2007social, walther1996computer}. It also provides a realistic setting for cross-cultural interaction: participants share an occupational context, but may differ in socio-demographic background, cultural region, and value system.

At each turn, participants were instructed to first infer their interlocutor's value system from the previous responses. They were then asked to reflect on whether the interlocutor's stance and underlying values aligned with their own, or whether they wanted the interlocutor to move toward their own stance and values. Based on this reflection, participants formulated their next response while continuing the discussion along one or more of the topic-specific aspects. This design encourages responses that are both conversationally grounded and value-aware, rather than isolated statements of opinion. It also aligns with prior work showing that persuasion and opinion change in online discussions depend not only on the content of arguments, but also on interactional dynamics, language choices, and the relationship between speakers' prior positions \cite{tan2016winning, durmus2019role}.

Participants were randomly paired across cultural backgrounds to ensure cross-cultural interaction. Each pair was also randomly assigned a \textbf{familiarity label}, with a 0.3 probability of familiarity, indicating whether the speakers were familiar with each other. We include familiarity because prior work suggests that social context and perceived relationship between interlocutors can shape politeness, disagreement, disclosure, and engagement in computer-mediated communication \cite{walther1996computer, danescu2013computational}. Thus, the familiarity label introduces a realistic source of conversational variation without deterministically tying it to participants' demographics or values.

The `Dialogue Prompt: Setup' and `Dialogue Prompt: Tasks' in appendix detail the prompts used to simulate each turn asynchronously from each participant in a dialogue. The number of turns in a conversation was sampled from a normal distribution with $\mu$=15 and $\sigma$=5. We use this stochastic length to approximate medium-length online discussions with sustained back-and-forth exchange, while avoiding a fixed conversation length that would make all simulations structurally identical. The stopping point was not revealed to participants. Once the sampled conversation length was reached, the discussion was forcefully concluded.

\begin{table}[h]
\centering
\renewcommand{\arraystretch}{1.1}
\setlength{\tabcolsep}{12pt}
\resizebox{\columnwidth}{!}{%
\begin{tabular}{cll}
\hline
\textbf{\#} & \textbf{Occupation} & \textbf{Topic}                        \\ \hline
1  & \multirow{3}{*}{\begin{tabular}[c]{@{}l@{}}Farm Workers,\\ Owners, Semi-, \& \\ Unskilled Workers\end{tabular}} & Water Usage Limits for Farmers          \\
2           &                     & Climate-Resilient Farming Practices   \\
3           &                     & Farmers' Access to Fair Pricing       \\ \cline{2-3} 
4  & \multirow{3}{*}{\begin{tabular}[c]{@{}l@{}}Professional,\\ Technical, \& \\Skilled Workers\end{tabular}}         & Local Ethical Standards for Global Tech \\
5           &                     & Remote Work Policy Changes            \\
6           &                     & AI-Driven Recruitment Systems         \\ \cline{2-3} 
7  & \multirow{3}{*}{\begin{tabular}[c]{@{}l@{}}Service, Sales, \&\\ Clerical Roles\end{tabular}}                   & Attendance Policies During Peak Seasons \\
8           &                     & Impact of Automation in Retail        \\
9           &                     & Employee Unionization in Retail       \\ \cline{2-3} 
10 & \multirow{3}{*}{Students}                                                                                       & AI-Driven Education and Assessment      \\
11          &                     & Cost of Higher Education              \\
12          &                     & Mental Health Support in Universities \\ \cline{2-3} 
13 & \multirow{3}{*}{\begin{tabular}[c]{@{}l@{}}Unemployed, \\ Retired, \& \\Housewife Roles\end{tabular}}             & Unemployment Benefits Program           \\
14          &                     & Financial Independence for Homemakers \\
15          &                     & Retirement Age Adjustments            \\ \hline
\end{tabular}
}
\caption{Occupation groups and their discussion topics.}
\vspace{-0.5cm}
\label{tab:topics}
\end{table}

% \begin{table}[t]
% \centering
% \small
% \renewcommand{\arraystretch}{1.1}
% \setlength{\tabcolsep}{12pt}
% \begin{tabular}{cl}
% \toprule
% \textbf{\#} & \textbf{Topic} \\
% \midrule
% 1  & AI-Driven Education and Assessment \\
% 2  & AI-Driven Recruitment Systems \\
% 3  & Attendance Policies During Peak Seasons \\
% 4  & Climate-Resilient Farming Practices \\
% 5  & Cost of Higher Education \\
% 6  & Employee Unionization in Retail \\
% 7  & Farmers’ Access to Fair Pricing \\
% 8  & Financial Independence for Homemakers \\
% 9  & Impact of Automation in Retail \\
% 10 & Local Ethical Standards for Global Tech \\
% 11 & Mental Health Support in Universities \\
% 12 & Remote Work Policy Changes \\
% 13 & Retirement Age Adjustments \\
% 14 & Unemployment Benefits Program \\
% 15 & Water Usage Limits for Farmers \\
% \bottomrule
% \end{tabular}
% \caption{List of value-laden discussion topics used in the WVS-grounded multi-agent simulation framework. \textbf{ADD OCCUPATION}}
% \label{tab:topics}
% \end{table}

% \subsection{Memory Update and Accumulative Value Change}
\subsection{Dialogue Summarization and Value Tracking}
Our simulation is longitudinal, where each persona participates in multiple dialogues with distinct interlocutors over time. To preserve continuity across these interactions, we maintain an independent memory for each participant. After every dialogue, we persist the participant's own turns, their reflection on their value system, and whether they intend to revise their values in future interactions. All subsequent conversations for that participant are conditioned on this memory, which is updated after each dialogue.

Storing such longitudinal interaction history is necessary for realistic simulations of social and value change. Prior work on generative agents shows that persistent memory and reflection are central to producing believable long-horizon behavior, since agents must be able to remember past experiences, synthesize them into higher-level reflections, and use them to guide future actions \cite{park2023generative}. Similarly, work on reflective language agents shows that textual reflections stored in episodic memory can improve later decision-making without modifying model weights \cite{shinn2023reflexion}. In our setting, this motivates treating each dialogue not as an isolated exchange, but as an experience that may incrementally shape the participant's later stance, reasoning, and conversational behavior.

However, naively persisting all prior turns word-for-word is impractical and undesirable, because: (i) longitudinal simulations produce a continually growing history that can exceed the model's context window. Prior work on LLM memory systems notes that limited context windows constrain extended conversations and motivate external or compressed memory mechanisms \cite{packer2024memgptllmsoperatingsystems}; (ii) raw transcript replay can introduce irrelevant details, redundancy, and topic drift into later prompts. We therefore summarize each completed dialogue into a structured memory record rather than storing the full transcript verbatim.

The structured summary contains: (i) the topic and main claim, (ii) the overall discussion trajectory, (iii) the agent's individual decisions, (iv) the detected stance change, and (v) the estimated short-term and long-term effects on both WVS dimensions. The \textit{Dialogue Summarizer Prompt} (in Appendix) details the prompt used to extract this information after each dialogue for every participant. This structured format serves two purposes. First, it compresses the dialogue into a form that can be included in future prompts under context-length constraints. Second, it preserves the aspects of the interaction that are most relevant to our research question: how repeated value-laden conversations may influence an agent's future beliefs and argumentative behavior.

The summary is appended to the agent's prior discussion history and included in subsequent prompts. Later conversations are therefore conditioned on earlier simulated experiences, enabling value change to accumulate incrementally rather than arising as a one-shot reaction to a single exchange. This design also makes the longitudinal process more interpretable: by storing both stance changes and estimated effects on the two WVS dimensions, we can trace how each dialogue contributes to the participant's evolving value profile over time.

% Since our simulation is longitudinal, where a persona engages in multiple dialogues with distinct people over time, after each dialogue, we persist each participant's turns and reflection about their own value systems and whether they want to change them in the future, into their independent memory. All future conversations of a participant are conditioned on this memory, which is subsequently updated after each conversation. However, since a conversation can be long, instead of persistent each turn word-by-word, we summarize them in a structured format, comprising the topic and main claim, the overall discussion trajectory, the agent's individual argumentative path, the detected stance change, and the estimated short-term and long-term effects on both WVS dimensions. The `Dialogue Summarizer Prompt' in Section \ref{appendix:prompts} details the prompt used to extract this information after each dialogue, for every participant. This summary is appended to the agent's prior discussion history and included in subsequent prompts. Later conversations are therefore conditioned on earlier simulated experiences, enabling value change to accumulate incrementally rather than arising as a one-shot reaction to a single exchange.
%Add a line on why we kept memory, why the conv was summarized, and why the value change was tracked after each conv
\subsection{Models}

We conduct simulations with three models, out of which two were API-based: GPT-4o\footnote{\url{https://openai.com/index/hello-gpt-4o/}} and Gemini-2.5-Flash\footnote{\url{https://deepmind.google/technologies/gemini/flash/}}, and one was a smaller open-weights model: Gemma-4-E4B-it\footnote{\url{https://ai.google.dev/gemma}}. All models are prompted identically using the same persona and discussion templates. Table \ref{tab:stats} summarizes key statistics. We incurred a total of $\sim$\$450 for the end-to-end simulation using GPT-4o, and $\sim$\$100 for Gemini-2.5-Flash. The open-weights Gemma-4-E4B-it model was run using one NVIDIA RTX 6000 GPU. 

\begin{table}[h]
\centering
\renewcommand{\arraystretch}{1}
\setlength{\tabcolsep}{10pt}
\begin{tabular*}{0.8\linewidth}{@{\extracolsep{\fill}}ll}
\toprule
\textbf{Metric} & \textbf{Value} \\
\midrule
Number of agents & 1,238 \\
Unique conversations & 3,895 \\
Number of topics & 15 \\
Avg. conversations per user & 6.16 \\
Avg. turns per conversation & 9.59 \\
\bottomrule
\end{tabular*}
% \caption{Statistics of the proposed WVS-grounded multi-agent simulation dataset, including the number of agents, generated conversations, discussion topics, and average interaction length across the longitudinal conversational setting.}
\caption{Statistics of the WVS-grounded simulation dataset}
\vspace{-0.5cm}
\label{tab:stats}
\end{table}

% ---------------------------------------------------------------
%  4. EVALUATION METHODOLOGY
% ---------------------------------------------------------------
\section{Evaluation Methodology \& Analysis}
% We evaluate the simulations along three questions. First, Are agents faithful to their assigned value systems? This evaluates how accurately models instantiate the intended cultural value profiles during interaction. Second, Do values drift over time? Where time refers to repeated interactions across multiple conversations, allowing us to measure whether agents’ value orientations drift throughout their conversational histories. Third, we measure how realistic are the conversations? This evaluates the stylistic and semantic diversity of generated discussions compared to human conversations. To automate the evaluation, we used two independent LLM evaluators, Gemini-3-Flash and GPT-5.2, alongside human annotators. We observed that human evaluators showed higher agreement with Gemini-3-Flash than with GPT-5.2. Consequently, we selected Gemini-3-Flash as the main evaluator. Further details are provided in ~\nameref{sec:eval-assess}.

We evaluate the simulations along three dimensions: \textbf{(i) Value Faithfulness:} Are agents faithful to their assigned value systems? This measures how accurately models instantiate the intended cultural value profiles during interaction. \textbf{(ii) Value Drift:} Do agents' values drift over time? Here, time refers to repeated interactions across multiple conversations, allowing us to measure whether agents' value orientations change over their conversational histories. \textbf{(iii) Conversational Realism:} How realistic are the generated conversations relative to human conversations? This evaluates the stylistic and semantic diversity of generated discussions compared to human dialogue. To automate the evaluation, we used two independent LLM evaluators, Gemini-3-Flash and GPT-5.2, alongside human annotators. We found that human evaluators showed higher agreement with Gemini-3-Flash than with GPT-5.2. Consequently, we selected Gemini-3-Flash as the main evaluator. Further details are provided in the Evaluator Assessment in the Appendix.

Because value attribution from dialogue is inherently subjective, we interpret the reported percentages as approximate estimates and focus our conclusions on qualitative and relative patterns. 

\subsection{Value Faithfulness}
% Value faithfulness measures whether a simulated persona correctly instantiates its assigned WVS value profile at the beginning of interaction. Specifically, we evaluate whether the value orientation reflected in an agent's first two conversations matches the values with which it was initialized. After each conversation, an LLM judge (Gemini-3-Flash) is prompted (using the Evaluation Prompt in the Appendix) to assess the agent's expressed value system along each WVS dimension. The judge uses a five-point ordinal scale: \textit{Strong Traditional (StrongTrad)/Leaning Traditional (LeanTrad)/Balanced/Leaning Secular (LeanSec)/Strong Secular (StrongSec)} for the Traditional--Secular-Rational axis, and \textit{Strong Survival (StrongSurv)/Leaning Survival (LeanSurv)/Balanced/Leaning Self-Expression (LeanSelf)/Strong Self-Expression (StrongSelf)} for the Survival--Self-Expression axis.

Value faithfulness measures whether a simulated persona instantiates its assigned WVS value profile at the beginning of interaction. We evaluate this using the agent's first two conversations, where accumulated memory is minimal and externally induced drift is least likely to have compounded. After each conversation, an LLM judge (Gemini-3-Flash) assessed the agent's expressed value system along each WVS dimension, using the Evaluation Prompt (in Appendix). The judge uses a five-point ordinal scale: \textit{Strong Traditional (StrongTrad)/Leaning Traditional (LeanTrad)/Balanced/Leaning Secular (LeanSec)/Strong Secular (StrongSec)} for the Traditional--Secular-Rational axis, and \textit{Strong Survival (StrongSurv)/Leaning Survival (LeanSurv)/Balanced/Leaning Self-Expression (LeanSelf)/Strong Self-Expression (StrongSelf)} for the Survival--Self-Expression axis.

We define two evaluation schemes. In the \textbf{Lenient} scheme, \textit{Strong}, \textit{Lean}, and \textit{Balanced} labels are treated as compatible with the assigned value pole. For example, a persona assigned a Traditional orientation is considered faithful if the judge returns \textit{StrongTrad}, \textit{LeanTrad}, or \textit{Balanced}. In the \textbf{Strict} scheme, only \textit{Lean} and \textit{Strong} labels consistent with the assigned orientation are accepted as faithful. Here, \textit{Balanced} is treated as a deviation because it indicates that the agent no longer expresses a clear directional commitment. Thus, the strict scheme measures unambiguous expression of the assigned value pole, while the lenient scheme captures broader compatibility with it.

% Faithfulness is assessed only over the first two conversations, when the agent’s accumulated memory is shortest and externally induced drift is least likely to have compounded, providing a clean window onto the model’s baseline value grounding. We define two evaluation schemes. In the \textbf{Lenient} scheme, \textit{Strong}, \textit{Lean}, and \textit{Balanced} labels are all collapsed toward the original value pole: a persona assigned a Traditional orientation is considered faithful if the judge returns any of \textit{StrongTrad}, \textit{LeanTrad}, or \textit{Balanced}. In the \textbf{Strict} scheme, only \textit{Lean} and \textit{Strong} labels consistent with the original orientation are accepted as faithful; \textit{Balanced} is treated as a deviation because it incorporates signals from the opposing value pole and indicates that the agent has lost a clear directional commitment.
% The strict scheme is intended to measure whether the assigned value pole remains clearly and unambiguously expressed in dialogue, whereas the lenient scheme evaluates broader compatibility with the assigned orientation.

\subsubsection{Are agents faithful to their initial value system?}

% Our experiments reveal that the agents simulated were not faithful to their initial value system. Table \ref{tab:faith} shows that across all evaluated models, fewer than 50\% of the personas remained faithful under the strict criterion, and even the lenient evaluation showed substantial failure rates. Gemma-4-E4B exhibited the highest unfaithfulness rates in both evaluation settings. GPT-4o and Gemini-2.5-Flash exhibit comparable unfaithfulness rates under the strict scheme, but differ slightly under the lenient scheme, where Gemini-2.5-Flash shows a lower faithfulness rate (47.5\%). These findings suggest that persona conditioning based on demographic and cultural information is frequently overridden by latent model priors acquired during pretraining and alignment.

Our experiments show that simulated agents often fail to remain faithful to their assigned value systems. As shown in Table~\ref{tab:faith}, across all evaluated models, fewer than 50\% of personas are faithful under the strict criterion, and even the lenient criterion shows substantial failure rates. Gemma-4-E4B exhibits the lowest faithfulness rates in both settings. GPT-4o and Gemini-2.5-Flash show comparable faithfulness rates under the strict scheme, but differ under the lenient scheme, where Gemini-2.5-Flash has a lower faithfulness rate (47.5\%). These findings suggest that persona conditioning based on demographic and cultural information is likely insufficient to override latent model priors acquired during pretraining and alignment.
While the reported percentages are based on automatic value attribution, we treat them as approximate estimates. Our main finding is the consistent qualitative asymmetry between initial value instantiation failures and subsequent value retention.
\begin{table}[h]
\centering
\small
\setlength{\tabcolsep}{10pt}
\begin{tabular}{lcc}
\toprule
 & Strict & Lenient \\
\midrule
GPT-4o              & 44.2\% & 52.5\% \\
Gemini-2.5-Flash    & 44.6\% & 47.5\% \\
Gemma-4-E4B         & 32.1\% & 37.1\% \\
\bottomrule
\end{tabular}
\caption{Strict and lenient faithfulness across models}
\vspace{-0.5cm}
\label{tab:faith}
\end{table}

\begin{figure*}[htbp]
    \centering
    \includegraphics[width=\textwidth, height=6cm]{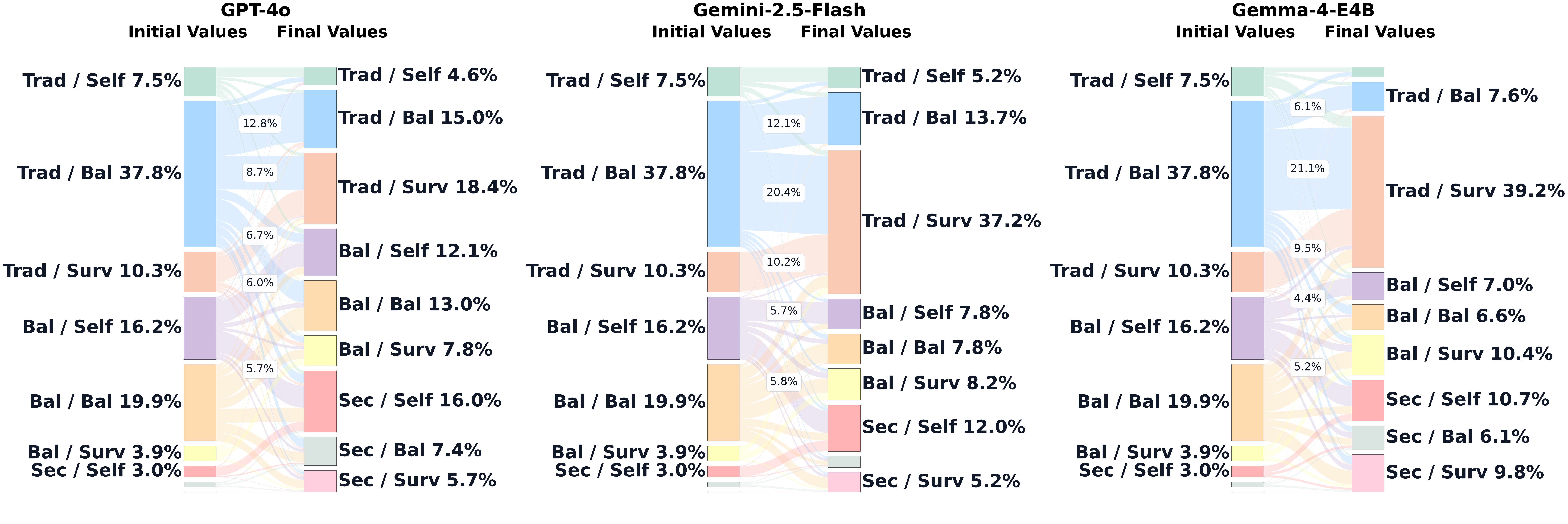}
    \caption{Value trajectories from initial WVS profile to final simulated value position across models}
    \vspace{-0.5cm}
    \label{fig:sankey}
\end{figure*}

\subsubsection{What drives value unfaithfulness?}
We fitted a binary logistic regression model using Statsmodels\footnote{\url{https://www.statsmodels.org/stable/index.html}} to identify factors associated with value faithfulness. The dependent variable was faithfulness, and the independent variables included demographic features (age, gender, country, region, education profile, employment status, economic status, immigration status, occupation, household status, ethnicity, and religion), the initial value system along both WVS dimensions, and communication style. We fit a separate model for each of the three evaluated LLMs. For each original feature, we use likelihood-ratio tests to assess whole-feature significance, while coefficient-level odds ratios ($OR$) indicate the direction and magnitude of specific category effects. The models were fitted on the full set of 1,200 agents. We did not use a train/test split because the goal was statistical inference rather than prediction. Since the outcome classes were approximately balanced (Table~\ref{tab:faith}), classification accuracy provides a coarse diagnostic of model fit. The three regressions achieved comparable accuracies of 72.1\%, 74.2\%, and 72.1\%, respectively.

% We fitted a binary logistic regression model (using Statsmodels\footnote{https://www.statsmodels.org/stable/index.html}) with the demographic features (age, gender, country, region, education profile, employment status, economic status, immigration status, occupation, household status, ethnicity, and religion), initial value system across both dimensions, and communication style as the independent variables and faithfulness as the dependent variable (outcome), for each of the three language models evaluated. For each original feature, we used likelihood-ratio tests to assess whole-feature significance, while coefficient-level odds ratios ($OR$) indicate the direction and magnitude of specific category effects. We fitted logistic regression on the full set of agents ($N= 1238$). We did not perform a train/test split because the objective was statistical inference rather than prediction. The outcome classes were approximately balanced \ref{tab:faith}. All three models achieved comparable classification accuracies of 72.1\%, 74.2\%, and 72.1\%, respectively.

\noindent
\textbf{Consistent predictors across all three models:} Three features were significant in all models: initial survival–self-expression dimension, initial traditional–secular-rational dimension, and communication style. Across all models, a persuasive communication style was the strongest and most consistent predictor of unfaithfulness (Gemini: $OR = 3.29$, $p < 0.001$; Gemma: $OR = 3.26$, $p < 0.001$; GPT-4o: $OR = 5.04$, $p < 0.001$). Conversely, initial value orientations on both WVS dimensions consistently predicted faithfulness: personas assigned a survival orientation (Gemini: $OR = 0.035$; Gemma: $OR = 0.045$; GPT-4o: $OR = 0.144$, all $p < 0.0014$), a secular-rational orientation (Gemini: $OR = 0.0734$; Gemma: $OR = 0.154$; GPT-4o: $OR = 0.212$, all $p < 0.001$), a traditional orientation (Gemini: $OR = 0.419$; Gemma: $OR = 0.467$; GPT-4o: $OR = 0.511$, all $p < 0.001$), and a self-expression orientation (Gemini: $OR = 0.323$; Gemma: $OR = 0.403$; GPT-4o: $OR = 0.554$, all $p < 0.001$) were all significantly more likely to remain faithful relative to their baselines. In particular, the protective effect of the orientation of initial values was strongest in Gemini and Gemma, and somewhat weaker in GPT-4o, suggesting that GPT-4o assigns less weight to the cultural profiles assigned when generating responses.

\noindent
\textbf{Model-specific predictors:} 
Beyond the shared predictors, each model exhibited unique associations. Under Gemini-2.5-Flash, economic status emerged as a significant feature (LR $p < 0.001$), with lower-class ($OR = 2.67$, $p < 0.001$) and working-class personas ($OR = 1.72$, $p = 0.019$) more likely to be unfaithful, and a both-low (low persuasive and low receptive) communication style also significantly associated with unfaithfulness ($OR = 2.20$, $p < 0.001$). Under Gemma-4-E4B, gender was a significant predictor (LR $p = 0.015$), with male personas more likely to be unfaithful ($OR = 1.42$,$ p = 0.013$); additionally, a receptive communication style was explicitly associated with faithfulness ($OR = 0.633$, $p = 0.015$). Under GPT-4o, cultural region reached significance (LR $p = 0.041$), with Latin American personas more likely to remain faithful ($OR = 0.600$, $p = 0.018$), and a both-low communication style also significantly predicted unfaithfulness ($OR = 2.13$, $p < 0.001$).

% Across all three models, communication style and initial value orientation are the primary determinants of faithfulness. 
We conjecture that persuasive communication prompts likely activate generic argumentative priors learned during alignment training, encouraging models to optimize for rhetorical effectiveness and social acceptability rather than strict adherence to assigned cultural value profiles. This suggests that conversational objectives and persona conditioning may compete during generation, with persuasive framing weakening recoverability of the intended value orientation.

\subsection{Value Drift}
% Value drift is measured as the shift between a persona's value position evaluated over its first two conversations and its position evaluated over its complete conversation history. The same five-point ordinal scale and Lenient/Strict evaluation schemes are applied, enabling change scores to be expressed as directional shifts toward or away from the initially assigned value pole. A persona is considered to have undergone a value change if its estimated overall position differs from its initial position (as reflected in the first two conversations) under the chosen scheme.
Value drift measures the shift between a persona's value position in its first two conversations and its value position over the complete conversation history. We use the same five-point ordinal scale and Lenient/Strict evaluation schemes as in the faithfulness analysis, allowing change scores to be expressed as directional shifts toward or away from the initially assigned value pole. A persona is considered to have undergone value drift if its estimated overall position differs from its initial position (as reflected in the first two conversations), under either of the chosen evaluation schemes.
\begin{table}[h]
\centering
\small
\setlength{\tabcolsep}{10pt}
\begin{tabular}{lcc}
\toprule
 & Strict & Lenient  \\
\midrule
GPT-4o              & 1.9\% & 0.8\% \\
Gemini-2.5-flash    & 3.9\% & 1\%   \\
Gemma-4-E4B         & 6.4\% & 2.4\% \\
\bottomrule
\end{tabular}
\caption{Strict and lenient value drift across models}
\vspace{-0.35cm}
\label{tab:val_change}
\end{table}

\subsubsection{Do values drift in simulated conversations?}

Table~\ref{tab:val_change} reports the percentage of personas exhibiting value drift for each simulated model. Across all models, drift rates remain low, particularly under the lenient criterion. GPT-4o shows the highest stability, with only 1.9\% strict and 0.8\% lenient value drift. Gemini-2.5-Flash exhibits slightly higher variation, but still maintains low drift overall. In contrast, Gemma-4-E4B shows the highest drift under both strict (6.4\%) and lenient (2.4\%) settings, indicating comparatively lower robustness in preserving value orientations over time. These results align with the stability of human values observed in real-world longitudinal research \cite{Bardi2009TheSO, vecchione2026stability}. Nonetheless, we do not claim that the simulated conversations correspond to the real-world temporal dynamics through which human values evolve. Rather, the simulations provide a controlled setting for evaluating whether models preserve value orientations across repeated interactions.

\subsubsection{Where do values land in a simulated society?}
For each model, Figure \ref{fig:sankey} shows the movement from the assigned initial value profile (on the left) to the final evaluated value profile (on the right), where the nodes represent combined positions of the two WVS cultural axes: Traditional--Secular-Rational and Survival--Self-Expression. The flow widths are proportional to the share of simulated users following that transition. Percent labels on the larger flows indicate the percentage of users in that model moving along that path. Each panel reports the total number of evaluated users and the share whose final value position differs from their initial profile.

Figure~\ref{fig:sankey} reveals a consistent pattern across models, where the largest transition occurs from Traditional/Balanced to Traditional/Survival. This flow accounts for 12.8\%, 20.5\%, and 21.2\% of personas in GPT-4o, Gemini-2.5-Flash, and Gemma-4-E4B, respectively. This suggests that personas initialized as balanced on the Survival--Self-Expression dimension are often evaluated as more Survival-oriented after repeated conversations. Correspondingly, the Traditional/Survival cell grows from 10.3\% at initialization to 18.4\% for GPT-4o, 37.2\% for Gemini-2.5-Flash, and 39.2\% for Gemma-4-E4B. The direction of this shift is consistent across models, differing mainly in magnitude, which suggests a systematic tendency in conversational LLM simulations to favor Survival-oriented framings on value-laden topics, while the Traditional--Secular-Rational axis is comparatively more stable.

The Balanced/Balanced cell, representing personas with moderate positions on both WVS dimensions, contracts in every model: from 19.8\% at initialization to 13.0\% in GPT-4o, 7.7\% in Gemini-2.5-Flash, and 6.6\% in Gemma-4-E4B. This indicates that models struggle to preserve moderate value profiles, with many such personas moving toward Traditional/Survival. For social simulation, this is a notable fidelity failure, since the real WVS distribution contains substantial mass in the moderate region.

Interestingly, less frequent value combinations in the original WVS-derived samples, such as Balanced/Survival, Secular/Self-Expression, Secular/Balanced, and Secular/Survival, become more prevalent in the final evaluated profiles across all models. Thus, the simulated value space is not merely compressed toward a single dominant region. Rather, it is also reshaped in ways that amplify several initially rare value combinations. Together, these results show that repeated LLM-based interactions can systematically distort the intended population-level value distribution, even when individual-level drift rates appear low.

% Interestingly, across all models, less frequent value combinations, such as balanced/survival, secular/self-expression, secular/balanced, and secular/survival that were sampled from the real world (WVS data), show a strong increase where more individuals have these combinations as their evaluated value system after several conversations. 

% To further investigate faithfulness and value change, we conduct an ablation study on a subset of users (N=82), sampled to cover all combinations of WVS dimensions. We replicate the same simulations without demographic conditioning, retaining only the initial value system.
\begin{figure*}[t]
    \centering
    \includegraphics[width=\textwidth,height=6cm]{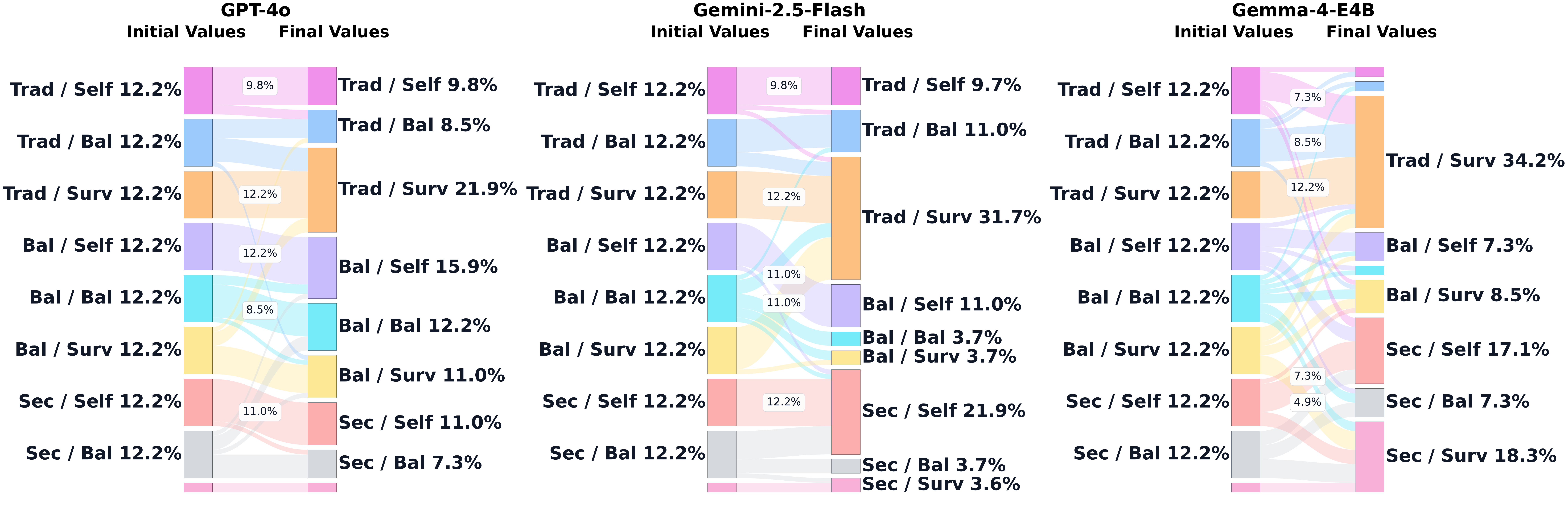}
    \caption{Value trajectories from initial WVS profile to final simulated value position without persona grounding}
    \vspace{-0.2cm}
    \label{fig:sankey-nopersona}
\end{figure*}

\subsection{Ablation: The Effect of Demographic Conditioning}
We further perform an ablation study in which we remove all demographic details from each user's persona and simulate conversations using only the assigned initial value system. This allows us to examine whether the observed patterns in value faithfulness and drift are driven by demographic persona conditioning, or whether they reflect model-level biases that persist even when demographic information is absent. The ablation is conducted on a subset of 82 users, sampled to uniformly cover all combinations of the WVS dimensions.

\begin{table}[ht]
\centering
\begin{tabular}{lcccc}
\toprule
& \multicolumn{2}{c}{Faithfulness} & \multicolumn{2}{c}{Value Drift} \\
\cline{2-3} \cline{4-5}
Model & Strict & Lenient & Strict & Lenient \\
\midrule
GPT-4o   & 71\%   & 72\%   & 3.7\% & 1.2\% \\
Gemini   & 68.3\% & 68.3\% & 4.9\% & 4.9\% \\
Gemma-4b & 39\%   & 42.7\% & 8.5\% & 3.7\% \\
\bottomrule
\end{tabular}
\caption{Ablation results: Faithfulness and value drift}
\vspace{-0.45cm}
\label{tab:no_persona_results}
\end{table}
\subsubsection{Value Faithfulness:} As detailed in Table \ref{tab:no_persona_results}, removing demographic details substantially improves value faithfulness for GPT-4o and Gemini-2.5-Flash. GPT-4o's strict faithfulness increases from 44.2\% to 71.0\% (+26.8), while Gemini-2.5-Flash increases from 44.6\% to 68.3\% (+23.7). For both models, the gap between the strict and lenient schemes also narrows, suggesting that their expressed values become more directionally consistent when only the value system is provided. Gemma-4-E4B improves more modestly, from 32.1\% to 39.0\% strict faithfulness (+6.9), and remains below 50\%. Overall, these results suggest that demographic conditioning can interfere with value instantiation, particularly for GPT-4o and Gemini-2.5-Flash, which appear better able to adhere to the assigned value system when demographic factors are excluded.

\subsubsection{Value Drift:} As detailed in Table \ref{tab:no_persona_results}, removing demographic factors slightly increases strict value drift for all models: GPT-4o rises from 1.9\% to 3.7\%, Gemini-2.5-Flash from 3.9\% to 4.9\%, and Gemma-4-E4B from 6.4\% to 8.5\%. Lenient drift rates remain low and broadly comparable to the full-persona condition, except for Gemini-2.5-Flash, where the strict and lenient rates both become 4.9\%. These results indicate that value drift remains limited even without demographic persona conditioning. Thus, the main effect of removing demographic information is not to eliminate longitudinal drift, but to improve initial value instantiation.

% As depicted in \textbf{[Table YYY]}, removing the demographic factors slightly increases the rate of strict value change for every model: GPT-4o rises from 1.9\% to 3.7\%, Gemini-2.5-Flash from 3.9\% to 4.9\%, and Gemma-4-E4B from 6.4\% to 8.5\%. Lenient rates remain low and broadly comparable to the persona condition, with the exception of Gemini, where the strict and lenient rates collapse to the same value (4.9\%). These results suggest that, even in the absence of persona conditioning, value change remains relatively stable across both evaluation settings.

\subsubsection{Value Landing:}
Figure \ref{fig:sankey-nopersona} shows the initial values distribution and the final evaluated values when simulating without demographic factors (persona). The dominant flow from Traditional/Balanced towards Traditional/Survival remains persistent across all models, similarly to when simulating with the demographic factors. The Balanced/Balanced collapse remains with Gemini and Gemma models, unlike GPT-4o, where the Balanced/Balanced distribution remains the same. 
The experiment also reveals that rare dimensions amplification significantly diminishes with GPT-4o and Gemini-2.5-Flash, only Gemma still strongly amplifies the Secular cells (Sec/Self: 12.2 to 17.1\%, Sec/Surv: 2.4 to 18.3\%). The overall observation is that GPT-4o is the most stable, and most starting cells largely stay put. Without demographic factors, GPT-4o behaves close to an identity function on the value space.

\subsection{Conversational Realism}

To assess the quality of simulated conversations, we evaluate two complementary aspects: \textbf{stylistic variation} and \textbf{content variability}. Stylistic variation captures whether conversations differ in how participants communicate, while content variability captures whether they differ in what semantic information they convey. For stylistic variation, we perform dialogue act tagging using the taxonomy of \citet{saha2021proto}, which comprises 11 tags, with Gemini-3-Flash as the tagger. For content variability, we compute semantic representations using 384-dimensional SentenceBERT embeddings \cite{reimers2019sentence}. We measure these properties along three axes:
\begin{itemize}
    \item \textbf{Individual style and content} measures inter-user similarity in terms of their conversational style and semantic content. For each user, we assign a dialogue act to each utterance across all conversations and aggregate these labels into a dialogue-act frequency vector representing the user's conversational style. We then compute pairwise cosine similarity across user pairs to quantify stylistic convergence. For content, we embed each utterance using SentenceBERT and compute a per-user mean embedding to represent the user's overall semantic profile.
    \item \textbf{Conversation-level diversity} measures how similar conversations are in terms of style and content. We aggregate dialogue-act and embedding-based representations at the conversation level, and compute pairwise distances between conversations to assess whether they are stylistically and semantically distinct.
    \item \textbf{Within-topic conversation diversity} measures whether conversations remain varied when the subject matter is held constant. For each topic, we compare style and content representations across different conversations on the same claim.
\end{itemize}
As a human baseline, we use the OUMdials dataset \cite{farag2022opening}, which contains human conversations on similarly open-ended and socially relevant topics. We apply the same dialogue-act and embedding-based analyses to OUMdials and compare the resulting style and content distributions with those from our simulated conversations.

\subsubsection{Individual-Level Diversity}
\begin{figure}[h]
    \centering
    \includegraphics[width=0.48\textwidth,height=4cm]{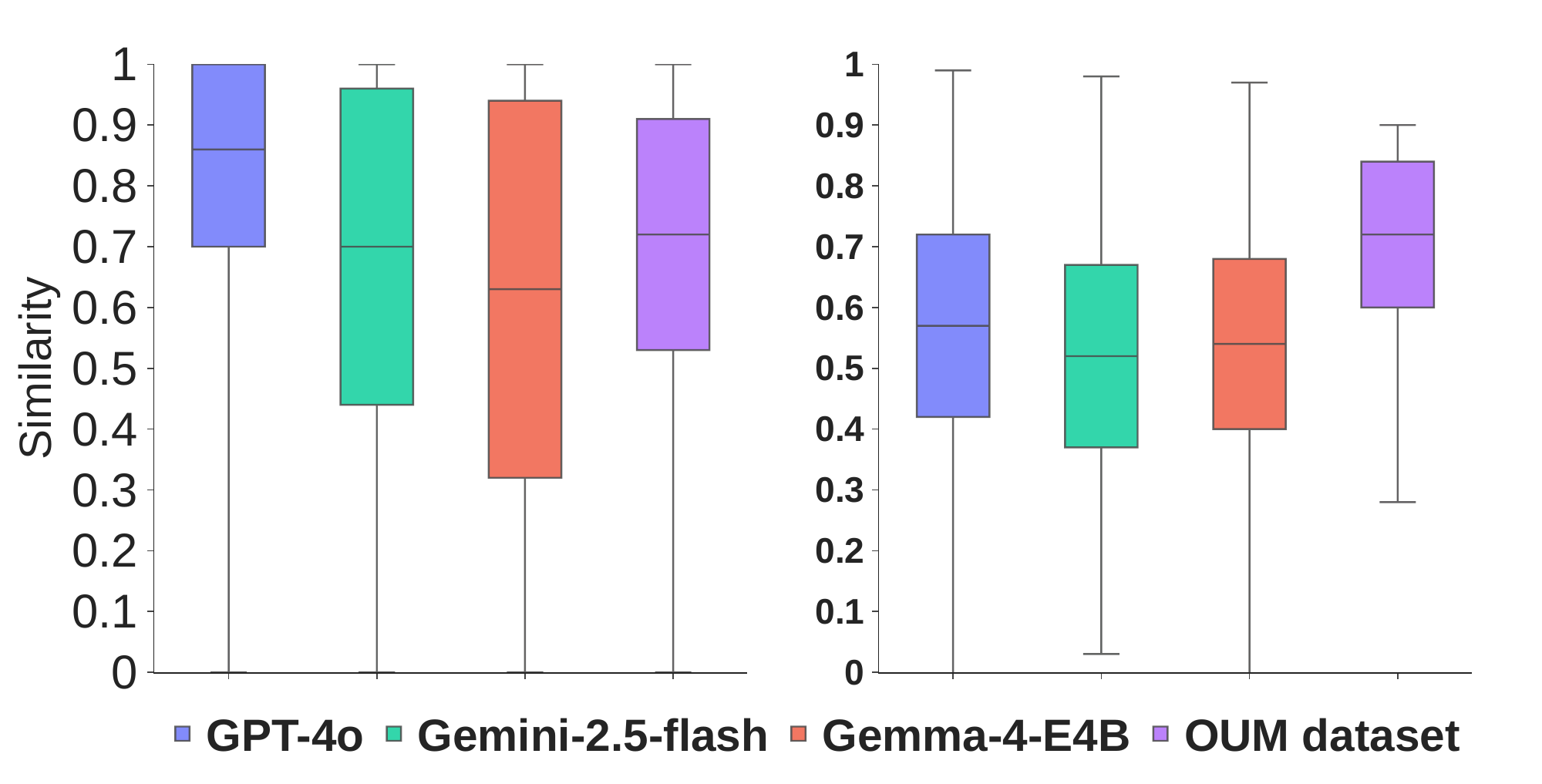}
    \caption{Individuals' style and content similarity}
    \vspace{-0.35cm}
    \label{fig:indiv_sim}
\end{figure}

Figure~\ref{fig:indiv_sim} presents box plots that capture individual-level similarity across our generated datasets, compared against the human-curated OUMdials baseline. %The box plots report the distribution of pairwise similarity scores across users, where the center line denotes the mean, the box denotes the standard deviation, and the whiskers denote the full range.
For \textbf{style}, GPT-4o exhibits the highest cross-user similarity, with a mean similarity of 0.86, exceeding the OUMdials baseline. This suggests that GPT-4o tends to produce stylistically homogeneous conversations across personas. Gemini-2.5-Flash overlaps closely with the OUMdials distribution, indicating a more human-like degree of stylistic variation. Gemma-4-E4B shows the greatest stylistic diversity, with a lower mean similarity of 0.63 and a broader distribution across user pairs.

For \textbf{content}, similarity scores are substantially lower than style scores across all generated datasets, with means below 0.6. This indicates that even when simulated users communicate in similar styles, they often differ in the semantic content they produce. In contrast, OUMdials shows the highest content similarity, with a mean of 0.73 and the tightest distribution. This suggests that human participants tend to discuss more similar content while expressing it through more varied styles, whereas LLM agents often display the opposite pattern: stronger stylistic convergence but more dispersed semantic content.
\begin{figure}[t]
    \centering
\includegraphics[width=0.48\textwidth,height=4cm]{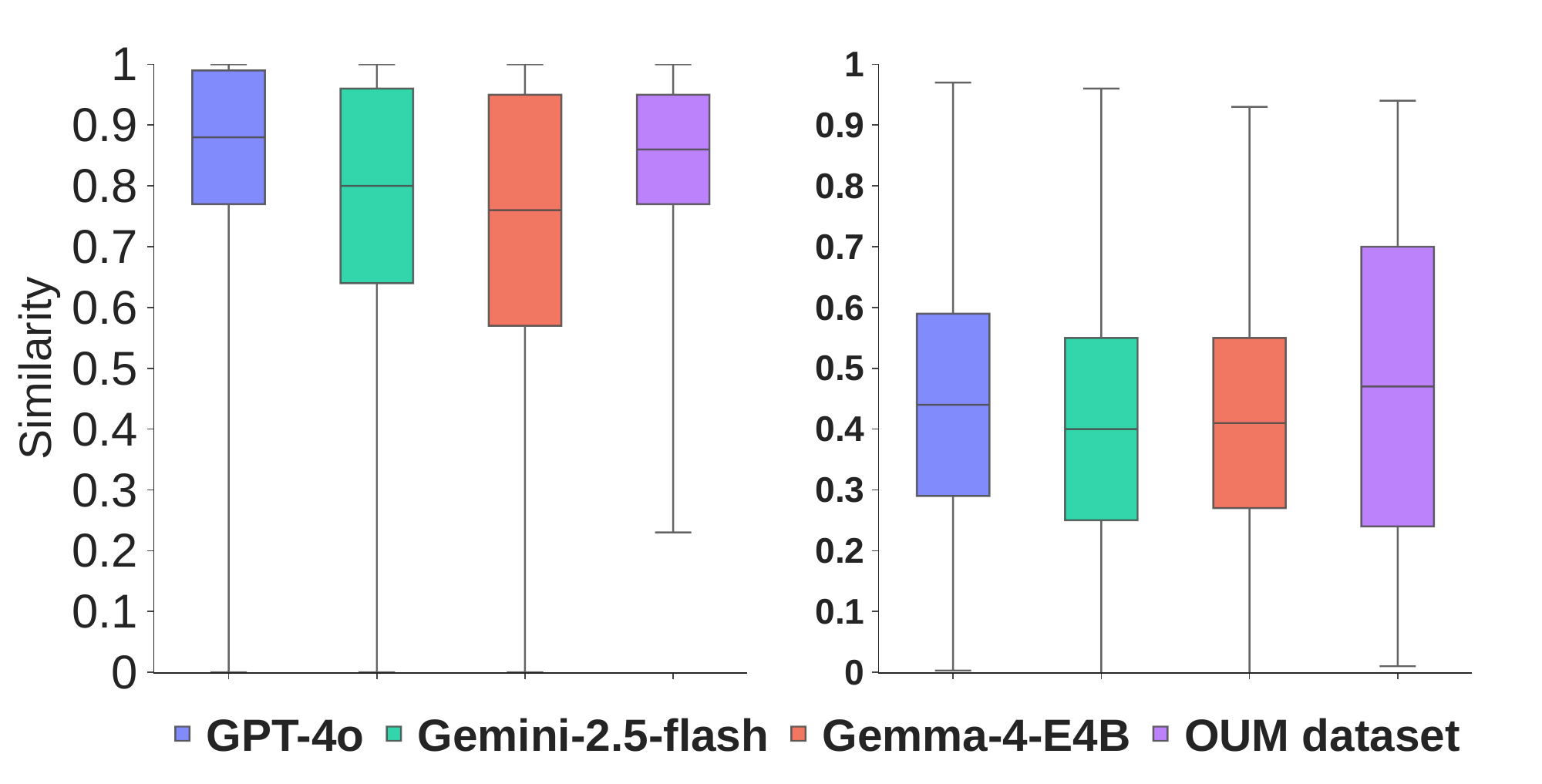}
    \caption{Conversations diversity by style and content}
    \vspace{-0.35cm}
    \label{fig:conv_sim}
\end{figure}

% Figure \ref{fig:indiv_sim} presents the individual-level similarity of conversations across our generated datasets compared to the human-curated OUMdials baseline. The box plots display the mean similarity (center line), the standard deviation (box), and the full range of similarity scores (whiskers). In terms of Style, GPT-4o exhibits the highest similarity across users, with a mean similarity (0.86) exceeding the OUMdials baseline, suggesting a tendency toward stylistically repetitive outputs across individuals. Gemini-2.5-Flash overlaps closely with the OUMdials distribution, while Gemma-4-E4B demonstrates the greatest stylistic diversity, evidenced by its lower mean (0.63) and broader distribution. In terms of Content, similarity scores are notably lower than stylistic scores across all generated datasets, with means falling below 0.6. This indicates that although the conversation style might be similar, individuals differ in their content. In contrast, the OUMdials dataset maintains the highest content consistency (0.73) with the tightest distribution, highlighting that humans discuss similar things in less similar ways.
\subsubsection{Conversation-Level Diversity}
% Figure \ref{fig:conv_sim} illustrates the similarity of conversations within our generated datasets compared to the human-curated OUMdials baseline. The box plots display the mean similarity (center line), the standard-deviation (box), and the full range of similarity scores (whiskers).

% In terms of Style, all models show significant overlap with the OUMdials dataset. However, Gemma-4-E4B exhibits the greatest diversity, evidenced by its lower median and broader distribution. 
% In terms of Content, similarity scores are notably lower than stylistic scores across all models and baselines, with means falling below $0.5$. This indicates that the datasets have high thematic variety. Specifically, Gemini-2.5-Flash and Gemma-4-E4B demonstrate the most varied content distributions, as their mean similarity scores are the lowest among them.

Figure~\ref{fig:conv_sim} compares conversation-level similarity in the generated datasets against the OUMdials baseline. For \textbf{style}, all models substantially overlap with OUMdials, suggesting that the overall distribution of dialogue acts is broadly comparable to human conversations at the exchange level. Gemma-4-E4B shows the greatest stylistic diversity, reflected by its lower median similarity and wider spread. For \textbf{content}, similarity scores are lower than stylistic scores across both generated and human datasets, with means below $0.5$. This indicates high thematic variation across conversations. Among the simulated models, Gemini-2.5-Flash and Gemma-4-E4B show the most diverse content distributions, with the lowest mean similarity scores.

\subsubsection{Within-Topic Conversation Diversity}
\begin{figure}[t]
    \centering
    \includegraphics[width=0.48\textwidth,height=4cm]{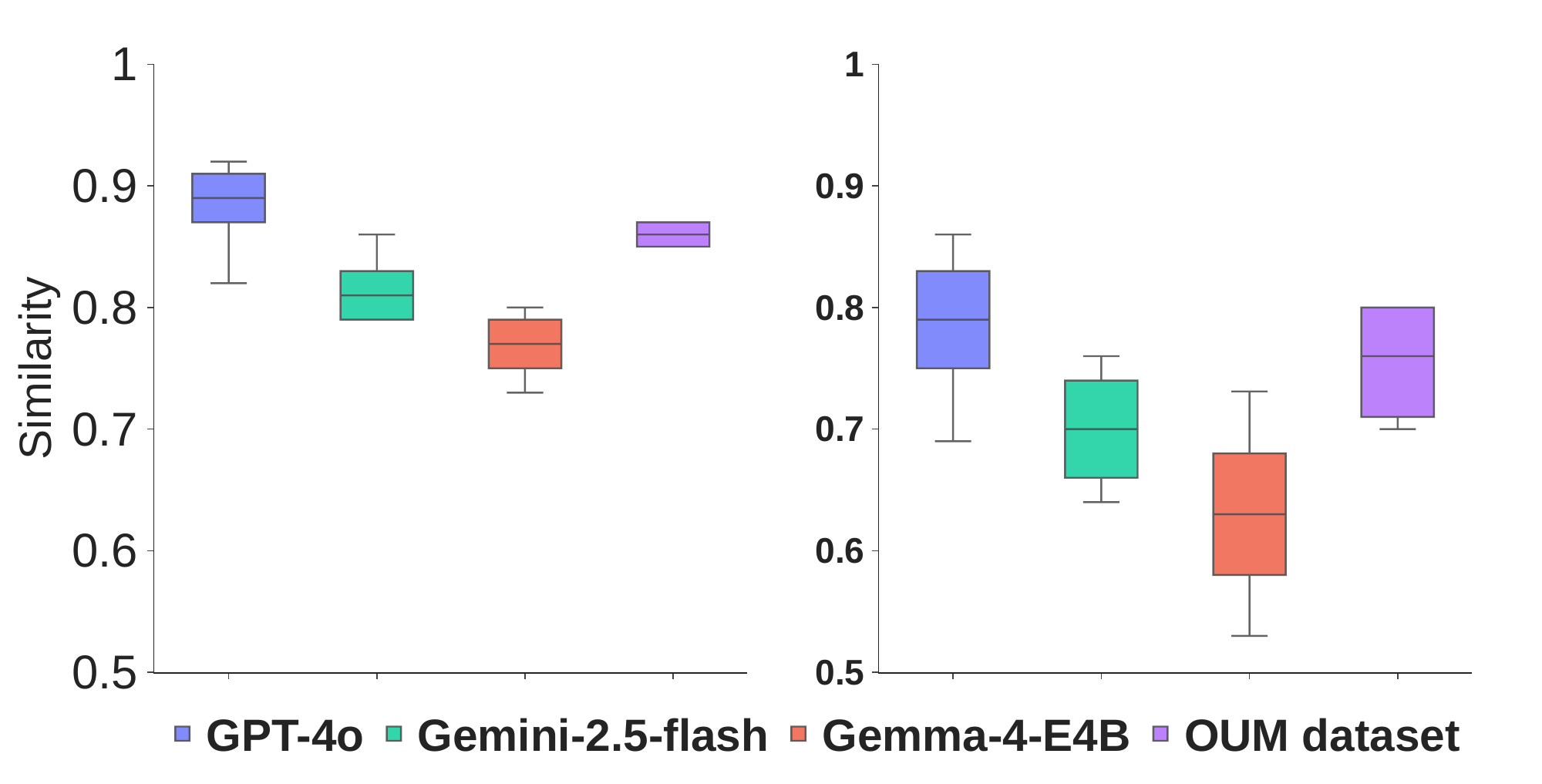}
    \caption{Within-topic style and content diversity}
    \label{fig:topic_conv}
    \vspace{-0.35cm}
\end{figure}

% Figure \ref{fig:topic_conv} illustrates the similarity of conversations grouped by topic. In terms of Style, Gemma-4-E4B demonstrates the highest diversity (lowest mean similarity), significantly exceeding the consistency levels of the human-curated OUMdials dataset. Conversely, GPT-4o maintains a highly uniform style across topics. Regarding Content, Gemma-4-E4B again shows the most varied output, with a mean similarity of  $0.63$ and a notably larger standard deviation compared to other sources. This suggests that Gemma-4-E4B explores a broader range of informational content within specific topics. In contrast, both GPT-4o and the OUMdials dataset exhibit higher similarity scores ($> 0.75$), indicating a more focused or repetitive content distribution within thematic clusters. We provide a further analysis on topic-level diversity in within-topic Conversations Diversity (in the Appendix). 
Figure~\ref{fig:topic_conv} shows conversation similarity when conversations are grouped by topic. For \textbf{style}, Gemma-4-E4B exhibits the greatest diversity, with the lowest mean similarity and a broader distribution than the OUMdials baseline. In contrast, GPT-4o maintains a highly uniform dialogue-act profile across conversations on the same topic, indicating stronger stylistic convergence. For \textbf{content}, Gemma-4-E4B again shows the greatest variation, with a mean similarity of $0.63$ and a substantially larger spread than the other sources. This suggests that Gemma-4-E4B explores a wider range of semantic content even when the underlying claim is held fixed. By contrast, both GPT-4o and OUMdials show higher within-topic content similarity ($>0.75$), indicating more focused discussions within thematic clusters. We provide additional topic-level analysis in the Appendix.

\section{Related Work}

\subsection{LLM Value Evaluation}
\label{related_work}
% Research on value alignment in LLMs has expanded considerably in the last years, methods emerged such as  ystematic evaluations spanning ethical frameworks, demographic variation, and cross-cultural differences \cite{cao2023assessing,tao2024cultural,santurkar2023whose}. 
Research on value alignment in LLMs has increased in recent years, with systematic evaluations emerging across ethical frameworks, demographic variation, and cross-cultural differences \cite{cao2023assessing,tao2024cultural,santurkar2023whose}.
Most existing approaches rely on established social-science instruments such as Schwartz's value taxonomy \cite{schwartz1999multimethod}, the WVS \cite{Inglehart_Welzel_2005}, or Hofstede's cultural dimensions \cite{hofstede2001culture}, and evaluate models through single-turn elicitation using Likert-scale questions or value inventories. Across these studies, a recurring finding is that LLMs tend to lean toward Western, educated, liberal orientations regardless of persona conditioning \cite{santurkar2023whose,johnson2022ghost,dwivedi-etal-2023-eticor}.
However, these evaluations remain fundamentally static. They measure what a model states when directly prompted, but not whether assigned value systems are behaviorally instantiated in interaction. Recent work has begun to expose this limitation. \citet{shen2025mind} shows that LLMs often exhibit a value-action gap, where stated values diverge from the choices models make in contextualized scenarios. Similarly, benchmarks such as ValueBench \citep{ren2024valuebench} attempt to move beyond isolated survey responses, yet still remain confined to single-turn settings. Existing evaluations provide limited visibility into whether assigned value profiles persist through sustained, multi-turn interaction, especially under disagreement and cross-cultural contestation. We address this gap by testing whether culturally grounded value profiles are faithfully instantiated and remain stable across repeated social interactions.

\subsection{Multi-Agent LLM Systems}
Multi-agent LLM simulations are increasingly used to model social behavior, including emergent group dynamics, opinion formation, and collective decision-making shaped by heterogeneous value systems \cite{park2023generative,chuang2024simulating,osman2023human}. Prior work shows that LLM societies can exhibit coordination, relationship formation, and information diffusion without explicit scripting \cite{park2024generative}, while value diversity can improve emergent intelligence and network integration \cite{osman2023human}. However, most evaluations focus on task-level outcomes such as reasoning, planning, coordination, or social intelligence benchmarks \cite{zhou2024sotopia}, rather than whether agents remain faithful to their assigned cultural identities during repeated interaction. Recent work such as ValueFlow \cite{liu2026valueflow} studies how value perturbations propagate through interacting LLM societies, but still assumes that agents initially instantiate stable value orientations. Our work examines this assumption directly by evaluating whether WVS-grounded personas both faithfully instantiate and preserve assigned value systems throughout longitudinal interaction. Moreover, our simulation relies on real personas drawn from the WVS data rather than synthetic personas, offering a closer approximation to the heterogeneity and internal consistency of real human value profiles than synthetically generated personas can provide.  

\subsection{Value Drift}
Psychology suggests that human values are generally stable over time \cite{Lane_1976}. Prior studies show that core value systems remain relatively persistent across adulthood and even adolescence, despite moderate shifts in specific priorities \cite{milfont2016values, vecchione2016stability, vecchione2026stability}. At the same time, values are also context-sensitive and can change through social interaction. Psychological research identifies both unconscious and deliberate mechanisms of value change, including priming, reflection, and consistency maintenance \cite{bardi2011dual}. Other work shows that values may be reconstructed differently when applied to concrete situations rather than abstract principles \cite{maio1998values, seligman2013dynamics}. These dynamics have also been explored in agent-based simulations studying moral change and collective policy behavior \cite{de2024modelling, dehkordi2024examining}.

% At the same time, psychological research also emphasizes that values are context-sensitive and can be reconstructed dynamically during social interaction. Empirical psychology identifies several mechanisms through which human values can drift. \citet{bardi2011dual} distinguish between a peripheral unconscious route (priming, identification) and a central-conscious route (deliberation, consistency maintenance). \citet{maio1998values} showed that because many values function as truisms, targeted ``inoculation'' challenges can trigger genuine reflection and value re-evaluation. Situational and contextual factors also matter, \citet{seligman2013dynamics} argued that abstract values appear stable but are reconstructed when applied to concrete issues. Agent-based models have formalized these dynamics at the population level, modeling how moral revolutions arise \cite{de2024modelling} and how openness to value change shapes collective policy responses \cite{dehkordi2024examining}. 

% ---------------------------------------------------------------
 % DISCUSSION AND BROADER IMPACT
% ---------------------------------------------------------------
\section{Discussion and Conclusion}
% Our findings have broader repercussions, although, on one hand, a growing body of work is interested in using LLMs as proxies for real users, works such as \cite{saha-choudhury-2025-user, pandey-etal-2025-generate} show that LLMs can't simulate individuals, and, at best operate at larger group-level. Hence, this questions if such prompt-based approaches using models like LLMs can enable digital twins and digital humanities. Many works look to create digital clone of humans or cells, or other aspects of society to simulate different interventions. However, if models are not faithful to start with, it raises questions on the utility of such methods, and if the resulting studies will be generalizable to all human population, or be applicable to a few.

While LLM-based simulations increasingly model individuals and populations, our results show that conversational plausibility can mask failures of representational fidelity. They may therefore reproduce long-standing external validity problems, including overgeneralization from WEIRD populations \citep{henrich2010weirdest}. These risks might get amplified in synthetic populations, digital twins, and policy simulations, where unfaithful value profiles can distort estimates of public opinion, intervention effects, or group behavior. Such distortions may especially affect marginalized or culturally underrepresented groups and contribute to algorithmic monoculture, where downstream studies inherit the same model-specific biases \citep{kleinberg2021algorithmic, saha-choudhury-2025-user, pandey-etal-2025-generate, saha-etal-2025-meta, saha-etal-2025-reading}. Also, since richer demographic prompting is not a sufficient safeguard, LLM-based simulations should be treated as virtual respondents, but as model-mediated instruments requiring validation against the populations they claim to represent \citep{bisbee2024synthetic,ziems2024can,madden2025evaluating}.

In conclusion, we presented a large-scale WVS-grounded framework for evaluating cross-cultural value faithfulness and drift in LLM-based multi-agent discussions. Under our evaluation protocol, current LLM agents generate coherent conversations but often fail to instantiate their assigned value profiles before exhibiting comparatively little measurable longitudinal drift. Because value attribution from dialogue is inherently subjective, the reported frequencies should be interpreted as approximate estimates rather than precise point values. Nevertheless, the qualitative asymmetry between initial instantiation failures and subsequent value drift is consistent across all evaluated models. These findings motivate evaluating LLM societies through behavioral consistency under interaction, rather than relying solely on static survey-style responses.

% In conclusion, we presented a large-scale WVS-grounded framework for evaluating cross-cultural value faithfulness and drift in LLM-based multi-agent discussions. Under our evaluation protocol, current LLM agents generate coherent conversations but frequently fail to instantiate their assigned value profiles before exhibiting measurable longitudinal drift. While the exact frequencies depend on automatic value attribution, this qualitative asymmetry is consistent across all evaluated models. This calls for evaluating LLM societies through behavioral consistency under interaction, rather than relying only on static survey-style responses.

% Our results show that current LLM agents can generate coherent conversations, but they do not reliably represent or preserve diverse human value profiles over time. This calls for evaluating LLM societies through behavioral consistency under interaction, rather than relying only on static survey-style responses.

\paragraph{Limitations.}
This work exhibits the following limitations. First, value orientations were evaluated using an LLM-based judge, which may reflect cultural or alignment-related biases despite moderate agreement with human annotators, indicating that value attribution is inherently subjective. Accordingly, our reported percentages should be interpreted as approximate measurements. Future work should validate the evaluation using multiple independent judges and larger human-annotated subsets.
Second, our simulations were conducted using only three language models, limiting the generalizability of the findings across broader open-source, multilingual models. Third, the simulated conversations were exclusively in English, which may suppress the cultural nuances and communication patterns present in multilingual real-world interactions. Finally, the simulations rely on synthetic conversational settings and predefined discussion topics, which may not fully capture the complexity of real-world social interaction.

\section*{Ethics Statement}
This work does not aim to recommend or prescribe which value systems societies should adopt. Rather, our goal is observational: to analyze how current LLM agents instantiate, preserve, and modify culturally grounded value profiles during social interaction.
In addition, the simulated personas are derived from anonymized and aggregated WVS demographic information and are not identifiable to any real individuals. Synthetic names and concise biographies were generated solely to support realistic conversational grounding.
Finally, annotators were warned that some conversations may contain sensitive arguments due to the nature of value-laden social and political discussions. Therefore, we had their consent beforehand.

\bibliography{aaai2026}

@article{maio1998values,
  title={Values as truisms: evidence and implications.},
  author={Maio, Gregory R and Olson, James M},
  journal={Journal of personality and social psychology},
  volume={74},
  number={2},
  pages={294},
  year={1998},
  publisher={American Psychological Association}
}

@article{bardi2011dual,
  title={The dual route to value change: Individual processes and cultural moderators},
  author={Bardi, Anat and Goodwin, Robin},
  journal={Journal of cross-cultural psychology},
  volume={42},
  number={2},
  pages={271--287},
  year={2011},
  publisher={Sage Publications Sage CA: Los Angeles, CA}
}

@incollection{seligman2013dynamics,
  title={The dynamics of value systems},
  author={Seligman, Clive and Katz, Albert N},
  booktitle={The psychology of values},
  pages={53--75},
  year={2013},
  publisher={Psychology Press}
}

@article{de2024modelling,
  title={Modelling value change: An exploratory approach},
  author={de Wildt, Tristan and van de Poel, Ibo},
  journal={Journal of Artificial Societies and Social Simulation},
  volume={27},
  number={1},
  year={2024},
  publisher={JASSS}
}

@article{dehkordi2024examining,
  title={Examining the interplay between national strategies and value change in the battle against COVID-19: An agent-based modelling inquiry},
  author={Dehkordi, Molood Ale Ebrahim and Melnyk, Anna and Herder, Paulien and Ghorbani, Amineh},
  journal={Journal of Artificial Societies and Social Simulation},
  volume={27},
  number={1},
  year={2024},
  publisher={JASSS}
}

@incollection{schwartz1999multimethod,
  title={Multimethod probes of basic human values},
  author={Schwartz, Shalom H and Lehmann, Arielle and Roccas, Sonia},
  booktitle={Social psychology and cultural context},
  pages={107--124},
  year={1999},
  publisher={SAGE Publications, Inc.}
}

@article{osman2023human,
  title={Human Values in Multiagent Systems},
  author={Osman, Nardine and d'Inverno, Mark},
  journal={arXiv preprint arXiv:2305.02739},
  year={2023}
}

@article{choi2026overstating,
  title={Overstating Attitudes, Ignoring Networks: LLM Biases in Simulating Misinformation Susceptibility},
  author={Choi, Eun Cheol and Young, Lindsay E and Ferrara, Emilio},
  journal={arXiv preprint arXiv:2602.04674},
  year={2026}
}

@article{tjuatja2024llms,
  title={Do llms exhibit human-like response biases? a case study in survey design},
  author={Tjuatja, Lindia and Chen, Valerie and Wu, Tongshuang and Talwalkwar, Ameet and Neubig, Graham},
  journal={Transactions of the Association for Computational Linguistics},
  volume={12},
  pages={1011--1026},
  year={2024},
  publisher={MIT Press 255 Main Street, 9th Floor, Cambridge, Massachusetts 02142, USA~…}
}

@article{park2024generative,
  title={Generative agent simulations of 1,000 people},
  author={Park, Joon Sung and Zou, Carolyn Q and Shaw, Aaron and Hill, Benjamin Mako and Cai, Carrie and Morris, Meredith Ringel and Willer, Robb and Liang, Percy and Bernstein, Michael S},
  journal={arXiv preprint arXiv:2411.10109},
  volume={52},
  year={2024}
}

@inproceedings{zhou2024sotopia,
  title={Sotopia: Interactive evaluation for social intelligence in language agents},
  author={Zhou, Xuhui and Zhu, Hao and Mathur, Leena and Zhang, Ruohong and Yu, Haofei and Qi, Zhengyang and Morency, Louis-Philippe and Bisk, Yonatan and Fried, Daniel and Neubig, Graham and others},
  booktitle={International Conference on Learning Representations},
  volume={2024},
  pages={40975--41019},
  year={2024}
}

@article{argyle2023out,
  title={Out of one, many: Using language models to simulate human samples},
  author={Argyle, Lisa P and Busby, Ethan C and Fulda, Nancy and Gubler, Joshua R and Rytting, Christopher and Wingate, David},
  journal={Political Analysis},
  volume={31},
  number={3},
  pages={337--351},
  year={2023},
  publisher={Cambridge University Press}
}

@article{wang2025evaluating,
  title={Evaluating the ability of large language models to emulate personality},
  author={Wang, Yilei and Zhao, Jiabao and Ones, Deniz S and He, Liang and Xu, Xin},
  journal={Scientific reports},
  volume={15},
  number={1},
  pages={519},
  year={2025},
  publisher={Nature Publishing Group UK London}
}

@misc{haerpfer2022world,
  title={World values survey: Round seven-country-pooled datafile version 5.0},
  author={Haerpfer, Christian and Inglehart, Ronald and Moreno, Alejandro and Welzel, Christian and Kizilova, Kseniya and Diez-Medrano, Jaime and Lagos, Marta and Norris, Pippa and Ponarin, Eduard and Puranen, Bjorn and others},
  year={2022},
  publisher={JD Systems Institute \& WVSA Secretariat. https://doi. org/10.14281/18241.20}
}

@inproceedings{ren2024valuebench,
  title={ValueBench: Towards comprehensively evaluating value orientations and understanding of large language models},
  author={Ren, Yuanyi and Ye, Haoran and Fang, Hanjun and Zhang, Xin and Song, Guojie},
  booktitle={Proceedings of the 62nd Annual Meeting of the Association for Computational Linguistics (Volume 1: Long Papers)},
  pages={2015--2040},
  year={2024}
}

@book{Inglehart_Welzel_2005, 
place={Cambridge}, title={Modernization, Cultural Change, and Democracy: The Human Development Sequence}, 
publisher={Cambridge University Press}, 
author={Inglehart, Ronald and Welzel, Christian}, 
year={2005}}

@inproceedings{santurkar2023whose,
  title={Whose opinions do language models reflect?},
  author={Santurkar, Shibani and Durmus, Esin and Ladhak, Faisal and Lee, Cinoo and Liang, Percy and Hashimoto, Tatsunori},
  booktitle={International conference on machine learning},
  pages={29971--30004},
  year={2023},
  organization={PMLR}
}

@article{johnson2022ghost,
  title={The Ghost in the Machine has an American accent: value conflict in GPT-3},
  author={Johnson, Rebecca L and Pistilli, Giada and Men{\'e}dez-Gonz{\'a}lez, Natalia and Duran, Leslye Denisse Dias and Panai, Enrico and Kalpokiene, Julija and Bertulfo, Donald Jay},
  journal={arXiv preprint arXiv:2203.07785},
  year={2022}
}

@inproceedings{reimers2019sentence,
  title={Sentence-bert: Sentence embeddings using siamese bert-networks},
  author={Reimers, Nils and Gurevych, Iryna},
  booktitle={Proceedings of the 2019 conference on empirical methods in natural language processing and the 9th international joint conference on natural language processing (EMNLP-IJCNLP)},
  pages={3982--3992},
  year={2019}
}

@article{saha2021proto,
  title={Proto: A neural cocktail for generating appealing conversations},
  author={Saha, Sougata and Das, Souvik and Soper, Elizabeth and Pacquetet, Erin and Srihari, Rohini K},
  journal={arXiv preprint arXiv:2109.02513},
  year={2021}
}

@inproceedings{farag2022opening,
  title={Opening up minds with argumentative dialogues},
  author={Farag, Youmna and Brand, Charlotte and Amidei, Jacopo and Piwek, Paul and Stafford, Tom and Stoyanchev, Svetlana and Vlachos, Andreas},
  booktitle={Findings of the Association for Computational Linguistics: EMNLP 2022},
  pages={4569--4582},
  year={2022}
}

@book{hofstede2001culture,
  title={Culture's consequences: Comparing values, behaviors, institutions and organizations across nations},
  author={Hofstede, Geert},
  year={2001},
  publisher={Sage publications}
}

@inproceedings{shen2025mind,
  title={Mind the Value-Action Gap: Do LLMs Act in Alignment with Their Values?},
  author={Shen, Hua and Clark, Nicholas and Mitra, Tanu},
  booktitle={Proceedings of the 2025 Conference on Empirical Methods in Natural Language Processing},
  pages={3097--3118},
  year={2025}
}

@article{liu2026valueflow,
  title={ValueFlow: Measuring the Propagation of Value Perturbations in Multi-Agent LLM Systems},
  author={Liu, Jinnuo and Liu, Chuke and Shen, Hua},
  journal={arXiv preprint arXiv:2602.08567},
  year={2026}
}

@inproceedings{cao2023assessing,
  title={Assessing cross-cultural alignment between ChatGPT and human societies: An empirical study},
  author={Cao, Yong and Zhou, Li and Lee, Seolhwa and Piqueras, Laura Cabello and Chen, Min and Hershcovich, Daniel},
  booktitle={Proceedings of the first workshop on cross-cultural considerations in NLP (C3NLP)},
  pages={53--67},
  year={2023}
}

@article{tao2024cultural,
  title={Cultural bias and cultural alignment of large language models},
  author={Tao, Yan and Viberg, Olga and Baker, Ryan S and Kizilcec, Ren{\'e} F},
  journal={PNAS nexus},
  volume={3},
  number={9},
  pages={pgae346},
  year={2024},
  publisher={Oxford University Press US}
}

@book{chuang2024simulating,
  title={Simulating Human Opinion Dynamics Using AI Agents and Large Language Models},
  author={Chuang, Yun-Shiuan},
  year={2024},
  publisher={The University of Wisconsin-Madison}
}

@article{milfont2016values,
  title={Values stability and change in adulthood: A 3-year longitudinal study of rank-order stability and mean-level differences},
  author={Milfont, Taciano L and Milojev, Petar and Sibley, Chris G},
  journal={Personality and Social Psychology Bulletin},
  volume={42},
  number={5},
  pages={572--588},
  year={2016},
  publisher={Sage Publications Sage CA: Los Angeles, CA}
}

@article{vecchione2016stability,
  title={Stability and change of basic personal values in early adulthood: An 8-year longitudinal study},
  author={Vecchione, Michele and Schwartz, Shalom and Alessandri, Guido and D{\"o}ring, Anna K and Castellani, Valeria and Caprara, Maria Giovanna},
  journal={Journal of Research in Personality},
  volume={63},
  pages={111--122},
  year={2016},
  publisher={Elsevier}
}

@article{vecchione2026stability,
  title={Stability and change of basic personal values in mid-to-late adolescence: A 4-year longitudinal study},
  author={Vecchione, Michele and Spagnolo, Giorgia and Daniel, Ella and Benish-Weisman, Maya and Bardi, Anat},
  journal={European Journal of Personality},
  volume={40},
  number={2},
  pages={351--368},
  year={2026},
  publisher={Sage Publications Sage UK: London, England}
}

@misc{WVSWave7,
  author = {Haerpfer, C. and Inglehart, R. and Moreno, A. and Welzel, C. and Kizilova, K. and Diez-Medrano, J. and Lagos, M. and Norris, P. and Ponarin, E. and Puranen, B. and et al.},
  publisher = {World Values Survey Association},
  title = {World Values Survey Wave 7 (2017-2022) Cross-National Data-Set},
  year = {2022},
  version = {v6.0.0},
  doi = {10.14281/18241.24},
  url = {https://www.worldvaluessurvey.org/WVSDocumentationWV7.jsp}
}

@article{minson2022receptiveness,
  title={Receptiveness to opposing views: Conceptualization and integrative review},
  author={Minson, Julia A and Chen, Frances S},
  journal={Personality and Social Psychology Review},
  volume={26},
  number={2},
  pages={93--111},
  year={2022},
  publisher={Sage Publications Sage CA: Los Angeles, CA}
}

@article{yeomans2020conversational,
  title={Conversational receptiveness: Improving engagement with opposing views},
  author={Yeomans, Michael and Minson, Julia and Collins, Hanne and Chen, Frances and Gino, Francesca},
  journal={Organizational Behavior and Human Decision Processes},
  volume={160},
  pages={131--148},
  year={2020},
  publisher={Elsevier}
}

@article{cacioppo1982need,
  title={The need for cognition.},
  author={Cacioppo, John T and Petty, Richard E},
  journal={Journal of personality and social psychology},
  volume={42},
  number={1},
  pages={116},
  year={1982},
  publisher={American Psychological Association}
}

@book{petty2012communication,
  title={Communication and persuasion: Central and peripheral routes to attitude change},
  author={Petty, Richard E and Cacioppo, John T},
  year={2012},
  publisher={Springer Science \& Business Media}
}

@article{gudykunst1996influence,
  title={The influence of cultural individualism-collectivism, self construals, and individual values on communication styles across cultures},
  author={Gudykunst, William B and Matsumoto, Yuko and Ting-Toomey, Stella and Nishida, Tsukasa and Kim, Kwangsu and Heyman, Sam},
  journal={Human communication research},
  volume={22},
  number={4},
  pages={510--543},
  year={1996},
  publisher={Oxford University Press}
}

@article{park2012individual,
  title={Individual and cultural variations in direct communication style},
  author={Park, Hee Sun and Levine, Timothy R and Weber, Rene and Lee, Hye Eun and Terra, Lucia I and Botero, Isabel C and Bessarabova, Elena and Guan, Xiaowen and Shearman, Sachiyo M and Wilson, Marc Stewart},
  journal={International Journal of Intercultural Relations},
  volume={36},
  number={2},
  pages={179--187},
  year={2012},
  publisher={Elsevier}
}

@inproceedings{tan2016winning,
  title={Winning arguments: Interaction dynamics and persuasion strategies in good-faith online discussions},
  author={Tan, Chenhao and Niculae, Vlad and Danescu-Niculescu-Mizil, Cristian and Lee, Lillian},
  booktitle={Proceedings of the 25th international conference on world wide web},
  pages={613--624},
  year={2016}
}

@inproceedings{hidey2017analyzing,
  title={Analyzing the semantic types of claims and premises in an online persuasive forum},
  author={Hidey, Christopher and Musi, Elena and Hwang, Alyssa and Muresan, Smaranda and McKeown, Kathleen},
  booktitle={Proceedings of the 4th Workshop on Argument Mining},
  pages={11--21},
  year={2017}
}

@article{gao2013designing,
  title={Designing asynchronous online discussion environments: Recent progress and possible future directions},
  author={Gao, Fei and Zhang, Tianyi and Franklin, Teresa},
  journal={British Journal of Educational Technology},
  volume={44},
  number={3},
  pages={469--483},
  year={2013},
  publisher={Wiley Online Library}
}

@book{hew2012student,
  title={Student participation in online discussions: Challenges, solutions, and future research},
  author={Hew, Khe Foon and Cheung, Wing Sum},
  year={2012},
  publisher={Springer Science \& Business Media}
}

@article{boyd2007social,
  title={Social network sites: Definition, history, and scholarship},
  author={Boyd, Danah M and Ellison, Nicole B},
  journal={Journal of computer-mediated Communication},
  volume={13},
  number={1},
  pages={210--230},
  year={2007},
  publisher={Wiley Online Library}
}

@article{walther1996computer,
  title={Computer-mediated communication: Impersonal, interpersonal, and hyperpersonal interaction},
  author={Walther, Joseph B},
  journal={Communication research},
  volume={23},
  number={1},
  pages={3--43},
  year={1996},
  publisher={Sage Publications London}
}

@inproceedings{danescu2013computational,
  title={A computational approach to politeness with application to social factors},
  author={Danescu-Niculescu-Mizil, Cristian and Sudhof, Moritz and Jurafsky, Dan and Leskovec, Jure and Potts, Christopher},
  booktitle={Proceedings of the 51st Annual Meeting of the Association for Computational Linguistics (Volume 1: Long Papers)},
  pages={250--259},
  year={2013}
}

@inproceedings{durmus2019role,
  title={The role of pragmatic and discourse context in determining argument impact},
  author={Durmus, Esin and Ladhak, Faisal and Cardie, Claire},
  booktitle={Proceedings of the 2019 Conference on Empirical Methods in Natural Language Processing and the 9th International Joint Conference on Natural Language Processing (EMNLP-IJCNLP)},
  pages={5668--5678},
  year={2019}
}

@inproceedings{park2023generative,
  title={Generative agents: Interactive simulacra of human behavior},
  author={Park, Joon Sung and O'Brien, Joseph and Cai, Carrie Jun and Morris, Meredith Ringel and Liang, Percy and Bernstein, Michael S},
  booktitle={Proceedings of the 36th annual acm symposium on user interface software and technology},
  pages={1--22},
  year={2023}
}

@article{shinn2023reflexion,
  title={Reflexion: Language agents with verbal reinforcement learning},
  author={Shinn, Noah and Cassano, Federico and Gopinath, Ashwin and Narasimhan, Karthik and Yao, Shunyu},
  journal={Advances in neural information processing systems},
  volume={36},
  pages={8634--8652},
  year={2023}
}

@misc{packer2024memgptllmsoperatingsystems,
      title={MemGPT: Towards LLMs as Operating Systems}, 
      author={Charles Packer and Sarah Wooders and Kevin Lin and Vivian Fang and Shishir G. Patil and Ion Stoica and Joseph E. Gonzalez},
      year={2024},
      eprint={2310.08560},
      archivePrefix={arXiv},
      primaryClass={cs.AI},
      url={https://arxiv.org/abs/2310.08560}, 
}

@inproceedings{dwivedi-etal-2023-eticor,
    title = "{E}ti{C}or: Corpus for Analyzing {LLM}s for Etiquettes",
    author = "Dwivedi, Ashutosh  and
      Lavania, Pradhyumna  and
      Modi, Ashutosh",
    editor = "Bouamor, Houda  and
      Pino, Juan  and
      Bali, Kalika",
    booktitle = "Proceedings of the 2023 Conference on Empirical Methods in Natural Language Processing",
    month = dec,
    year = "2023",
    address = "Singapore",
    publisher = "Association for Computational Linguistics",
    url = "https://aclanthology.org/2023.emnlp-main.428/",
    doi = "10.18653/v1/2023.emnlp-main.428",
    pages = "6921--6931"
}

@article{stolcke-etal-2000-dialogue,
    title = "Dialogue act modeling for automatic tagging and recognition of conversational speech",
    author = "Stolcke, Andreas  and
      Ries, Klaus  and
      Coccaro, Noah  and
      Shriberg, Elizabeth  and
      Bates, Rebecca  and
      Jurafsky, Daniel  and
      Taylor, Paul  and
      Martin, Rachel  and
      Van Ess-Dykema, Carol  and
      Meteer, Marie",
    journal = "Computational Linguistics",
    volume = "26",
    number = "3",
    year = "2000",
    address = "Cambridge, MA",
    publisher = "MIT Press",
    url = "https://aclanthology.org/J00-3003/",
    pages = "339--374"
}

@inproceedings{DBLP:journals/corr/abs-1301-3781,
  author       = {Tom{\'{a}}s Mikolov and
                  Kai Chen and
                  Greg Corrado and
                  Jeffrey Dean},
  editor       = {Yoshua Bengio and
                  Yann LeCun},
  title        = {Efficient Estimation of Word Representations in Vector Space},
  booktitle    = {1st International Conference on Learning Representations, {ICLR} 2013,
                  Scottsdale, Arizona, USA, May 2-4, 2013, Workshop Track Proceedings},
  year         = {2013},
  url          = {http://arxiv.org/abs/1301.3781},
  bibsource    = {dblp computer science bibliography, https://dblp.org}
}

@inbook{inbook,
author = {Schwartz, Shalom},
year = {1992},
month = {12},
pages = {1-65},
title = {Universals in the Content and Structure of Values: Theoretical Advances and Empirical Tests in 20 Countries},
volume = {25},
isbn = {9780120152254},
journal = {Advances in Experimental Social Psychology},
doi = {10.1016/S0065-2601(08)60281-6}
}

@article{Lane_1976, title={The Nature of Human Values. By Milton Rokeach. (New York: The Free Press, 1973. Pp. 438. \$13.95.)}, volume={70}, DOI={10.2307/1959882}, number={3}, journal={American Political Science Review}, author={Lane, Robert E.}, year={1976}, pages={965–966}}

@article{Bardi2009TheSO,
  title={The structure of intraindividual value change.},
  author={Anat Bardi and Julie Anne Lee and Nadi Hofmann-Towfigh and Geoffrey Norman Soutar},
  journal={Journal of personality and social psychology},
  year={2009},
  volume={97 5},
  pages={
          913-29
        },
  url={https://api.semanticscholar.org/CorpusID:207726174}
}

@inproceedings{saha-choudhury-2025-user,
    title = "User Behavior Prediction as a Generic, Robust, Scalable, and Low-Cost Evaluation Strategy for Estimating Generalization in {LLM}s",
    author = "Saha, Sougata  and
      Choudhury, Monojit",
    editor = "Che, Wanxiang  and
      Nabende, Joyce  and
      Shutova, Ekaterina  and
      Pilehvar, Mohammad Taher",
    booktitle = "Findings of the Association for Computational Linguistics: ACL 2025",
    month = jul,
    year = "2025",
    address = "Vienna, Austria",
    publisher = "Association for Computational Linguistics",
    url = "https://aclanthology.org/2025.findings-acl.576/",
    doi = "10.18653/v1/2025.findings-acl.576",
    pages = "11047--11065",
    ISBN = "979-8-89176-256-5"
}

@inproceedings{pandey-etal-2025-generate,
    title = "To Generate or Discriminate? Methodological Considerations for Measuring Cultural Alignment in {LLM}s",
    author = "Pandey, Saurabh Kumar  and
      Saha, Sougata  and
      Choudhury, Monojit",
    editor = "Inui, Kentaro  and
      Sakti, Sakriani  and
      Wang, Haofen  and
      Wong, Derek F.  and
      Bhattacharyya, Pushpak  and
      Banerjee, Biplab  and
      Ekbal, Asif  and
      Chakraborty, Tanmoy  and
      Singh, Dhirendra Pratap",
    booktitle = "Proceedings of the 14th International Joint Conference on Natural Language Processing and the 4th Conference of the Asia-Pacific Chapter of the Association for Computational Linguistics",
    month = dec,
    year = "2025",
    address = "Mumbai, India",
    publisher = "The Asian Federation of Natural Language Processing and The Association for Computational Linguistics",
    url = "https://aclanthology.org/2025.findings-ijcnlp.95/",
    doi = "10.18653/v1/2025.findings-ijcnlp.95",
    pages = "1545--1562",
    ISBN = "979-8-89176-303-6"
}

@inproceedings{saha-etal-2025-meta,
    title = "Meta-Cultural Competence: Climbing the Right Hill of Cultural Awareness",
    author = "Saha, Sougata  and
      Pandey, Saurabh Kumar  and
      Choudhury, Monojit",
    editor = "Chiruzzo, Luis  and
      Ritter, Alan  and
      Wang, Lu",
    booktitle = "Proceedings of the 2025 Conference of the Nations of the Americas Chapter of the Association for Computational Linguistics: Human Language Technologies (Volume 1: Long Papers)",
    month = apr,
    year = "2025",
    address = "Albuquerque, New Mexico",
    publisher = "Association for Computational Linguistics",
    url = "https://aclanthology.org/2025.naacl-long.408/",
    doi = "10.18653/v1/2025.naacl-long.408",
    pages = "8025--8042",
    ISBN = "979-8-89176-189-6"
}

@article{henrich2010weirdest,
  title={The weirdest people in the world?},
  author={Henrich, Joseph and Heine, Steven J and Norenzayan, Ara},
  journal={Behavioral and brain sciences},
  volume={33},
  number={2-3},
  pages={61--83},
  year={2010},
  publisher={Cambridge University Press}
}

@article{kleinberg2021algorithmic,
  title={Algorithmic monoculture and social welfare},
  author={Kleinberg, Jon and Raghavan, Manish},
  journal={Proceedings of the National Academy of Sciences},
  volume={118},
  number={22},
  pages={e2018340118},
  year={2021},
  publisher={National Academy of Sciences}
}

@inproceedings{saha-etal-2025-reading,
    title = "Reading between the Lines: Can {LLM}s Identify Cross-Cultural Communication Gaps?",
    author = "Saha, Sougata  and
      Pandey, Saurabh Kumar  and
      Gupta, Harshit  and
      Choudhury, Monojit",
    editor = "Chiruzzo, Luis  and
      Ritter, Alan  and
      Wang, Lu",
    booktitle = "Proceedings of the 2025 Conference of the Nations of the Americas Chapter of the Association for Computational Linguistics: Human Language Technologies (Volume 1: Long Papers)",
    month = apr,
    year = "2025",
    address = "Albuquerque, New Mexico",
    publisher = "Association for Computational Linguistics",
    url = "https://aclanthology.org/2025.naacl-long.409/",
    doi = "10.18653/v1/2025.naacl-long.409",
    pages = "8043--8067",
    ISBN = "979-8-89176-189-6"
}

@article{bisbee2024synthetic,
  title={Synthetic replacements for human survey data? The perils of large language models},
  author={Bisbee, James and Clinton, Joshua D and Dorff, Cassy and Kenkel, Brenton and Larson, Jennifer M},
  journal={Political Analysis},
  volume={32},
  number={4},
  pages={401--416},
  year={2024},
  publisher={Cambridge University Press}
}

@article{ziems2024can,
  title={Can large language models transform computational social science?},
  author={Ziems, Caleb and Held, William and Shaikh, Omar and Chen, Jiaao and Zhang, Zhehao and Yang, Diyi},
  journal={Computational Linguistics},
  volume={50},
  number={1},
  pages={237--291},
  year={2024}
}

@article{madden2025evaluating,
  title={Evaluating the Use of Large Language Models as Synthetic Social Agents in Social Science Research},
  author={Madden, Emma Rose},
  journal={Journal of Social Computing},
  volume={6},
  number={4},
  pages={334--341},
  year={2025},
  publisher={TUP}
}
\appendix 
\clearpage
\section{Conversational Quality}
\subsection{Within-topic individual diversity} 
For each topic, we compare style and content profiles of different individuals discussing the same claim to assess whether cultural background produces meaningful variation even when the subject matter is held constant.

\begin{figure}[t]
    \centering
    \includegraphics[width=0.49\textwidth,height=6cm]{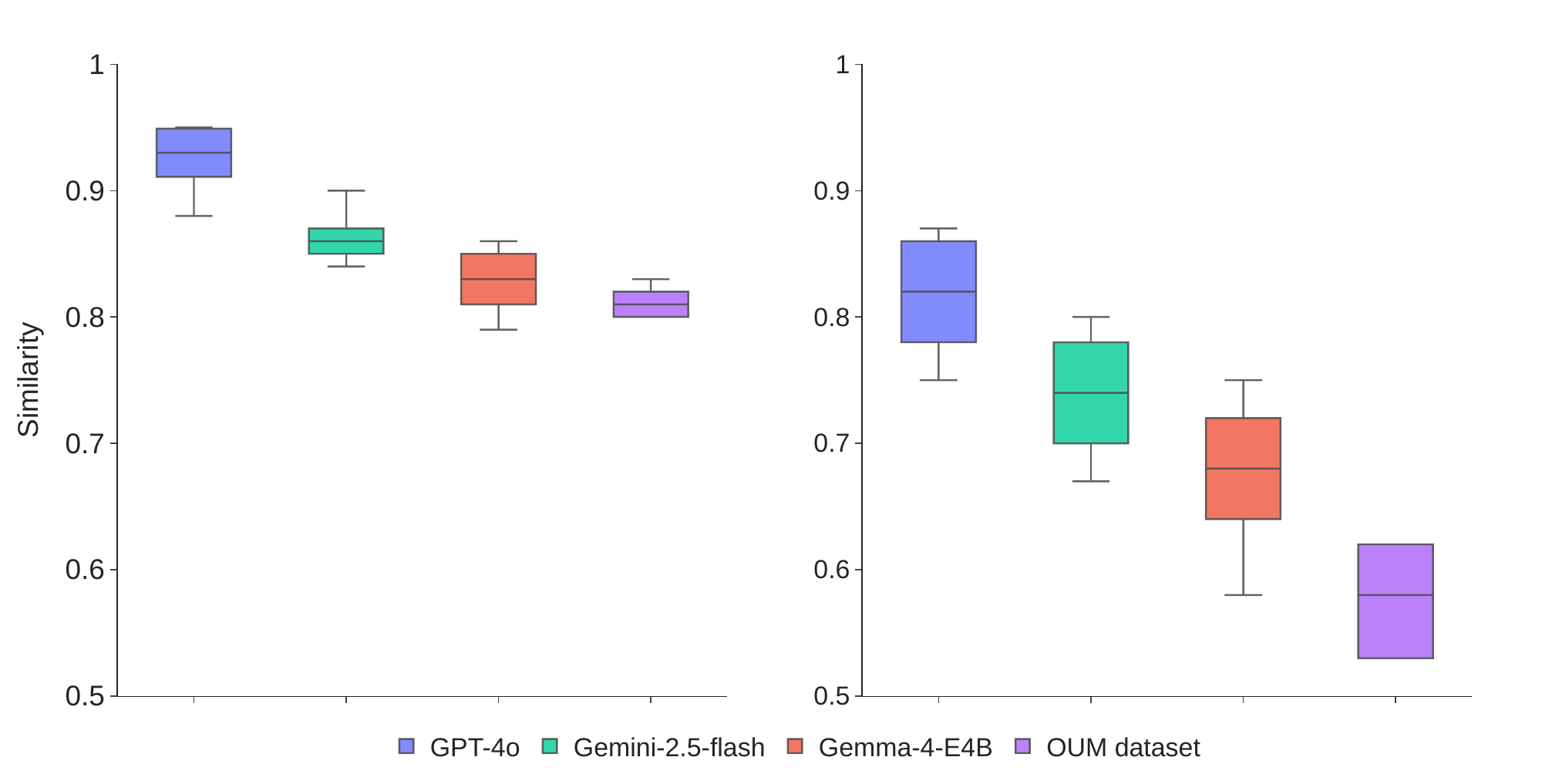}
    \caption{Mean Person-Topic Similarity.}
    \label{fig:topic_sim}
\end{figure}

% \subsubsection{Within-topic Individual's Diversity}
We report the mean per-person conversation similarity per topic averaged across all individuals. Figure \ref{fig:topic_sim} reveals that all AI models exhibit higher internal consistency compared to the human baseline in both style and content. GPT-4o shows the highest mean similarity (lowest diversity) in both categories, suggesting a tendency toward repetitive outputs when constrained by topic. In contrast, the OUM dataset displays significantly lower similarity, particularly in content (mean $\approx$ 0.58), highlighting that humans provide more varied information across different interactions on the same subject. Among the models, Gemma-4-E4B most closely approximates the human baseline's diversity levels, demonstrating a greater stylistic and thematic range than its counterparts.

\subsection{Within-topic Conversation Diversity}
\begin{figure*}[hptb]
    \centering
    \includegraphics[width=\textwidth,height=9.5cm]{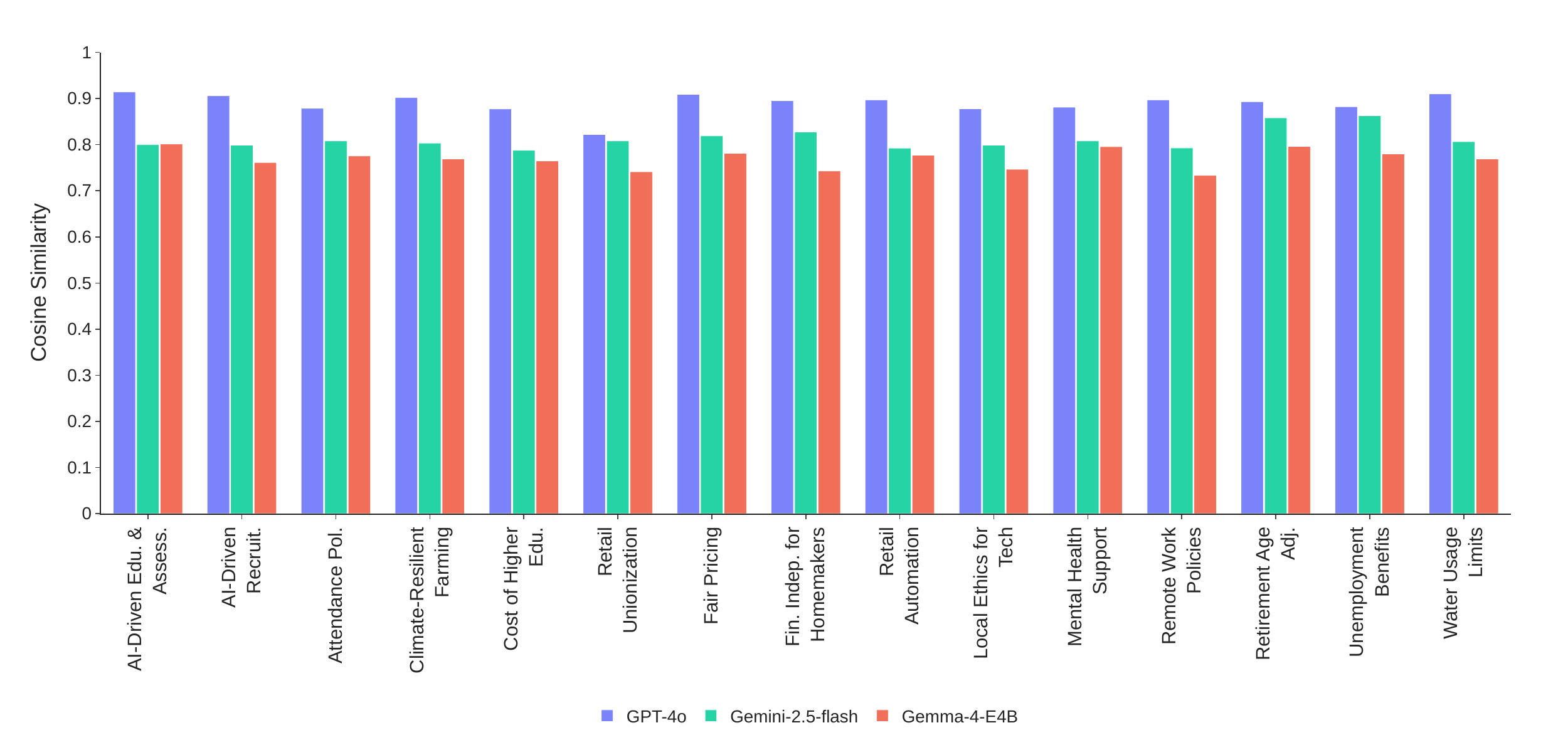}
    \caption{\textbf{Topic similarities - Style. }Topic-wise cosine similarity scores measuring stylistic consistency across GPT-4o, Gemini-2.5-flash, and Gemma-4-E4B. Higher similarity values indicate stronger preservation of stylistic alignment across generated discussions, with GPT-4o generally showing the highest consistency across most topics.}
        \vspace{-0.5cm}
    \label{fig:topic_sim_bar}
\end{figure*}

\begin{figure*}[hptb]
    \centering
\includegraphics[width=\textwidth,height=8cm]{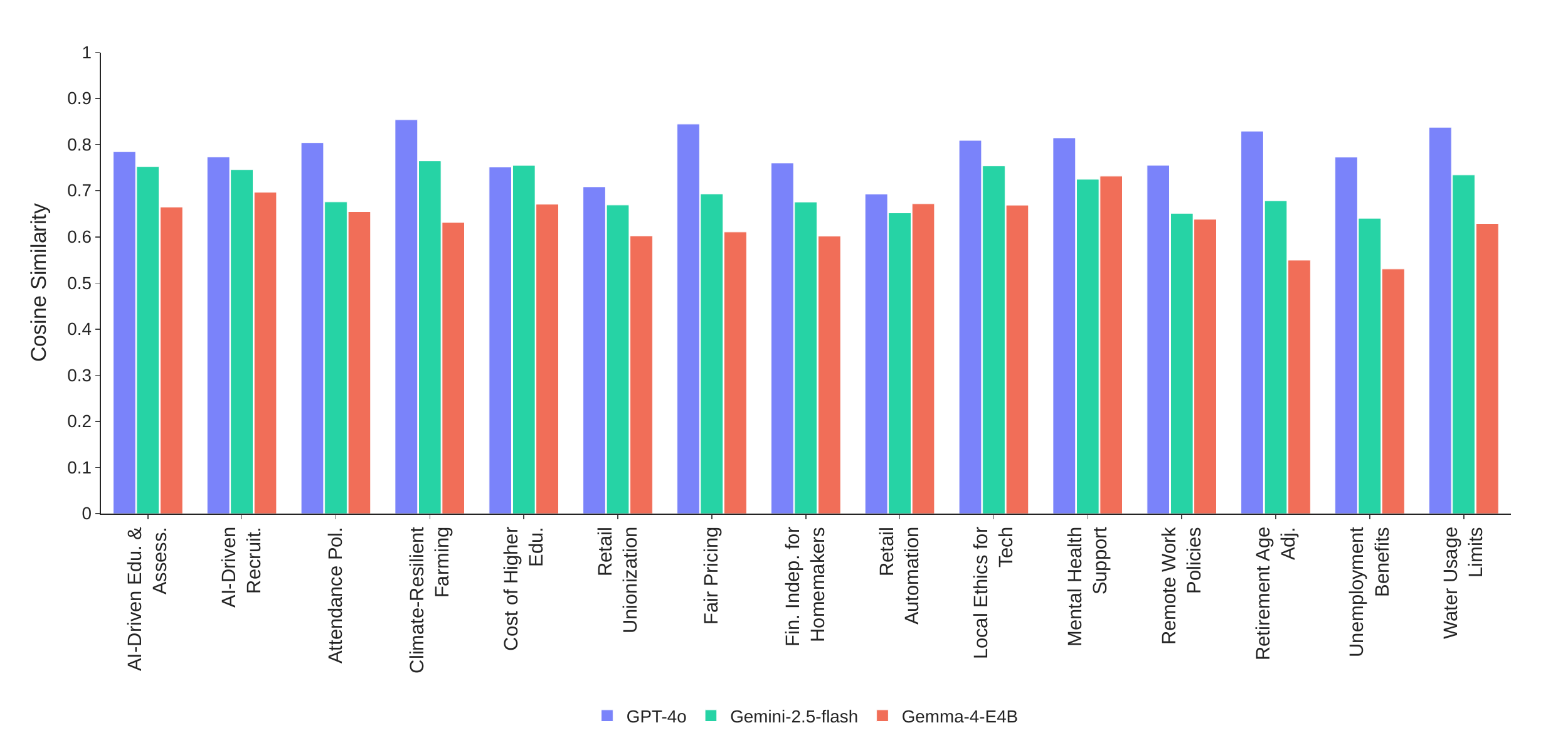}
    \caption{\textbf{Topic similarities - Content.} Topic-level cosine similarity scores comparing the semantic content consistency of simulated conversations generated by GPT-4o, Gemini-2.5-flash, and Gemma-4-E4B. The results show that GPT-4o generally maintains higher semantic alignment across topics, while smaller variations across models highlight differences in content diversity and consistency.}   
    \vspace{-0.5cm}
    \label{fig:topic_sim_bar_content}
\end{figure*}

We investigate the models' performance across diverse subject matter, Figures \ref{fig:topic_sim_bar} and \ref{fig:topic_sim_bar_content} present the mean cosine similarity for 15 topics across both content and style (dialog acts).In terms of Content, GPT-4o consistently exhibits the highest similarity scores across nearly all categories, peaking in topics such as "Climate-Resilient Farming Practices" and "Farmers' Access to Fair Pricing" (similarity $> 0.85$). This suggests a high degree of informational redundancy when the model addresses these themes. Conversely, Gemma-4-E4B demonstrates superior content diversity across the board, with particularly high variety in "Unemployment Benefits Program" and "Retirement Age Adjustments" where similarity scores drop significantly below $0.55$.Regarding Style (Dialog Acts), the models follow a similar tendency but with higher overall baseline similarity ($> 0.7$). GPT-4o maintains a very rigid stylistic structure regardless of the topic, while Gemma-4-E4B and Gemini-2.5-flash show greater flexibility. Notably, the stylistic gap between models is more uniform across topics than the content gap, indicating that while models may vary their information based on the subject, their conversational style remains relatively fixed.


% \begin{table*}[hptb]
% \centering
% \resizebox{\textwidth}{!}{%
% \begin{tabular}{llllll}
% \toprule
% \textbf{Abbrev.} & \textbf{Topic} &
% \textbf{Abbrev.} & \textbf{Topic} &
% \textbf{Abbrev.} & \textbf{Topic} \\
% \midrule

% AI-Driven Edu.\ \& Assess. & AI-Driven Education and Assessment &
% AI-Driven Recruit. & AI-Driven Recruitment Systems &
% Attendance Pol. & Attendance Policies During Peak Seasons \\

% Climate-Resilient Farming & Climate-Resilient Farming Practices &
% Cost of Higher Edu. & Cost of Higher Education &
% Retail Unionization & Employee Unionization in Retail \\

% Fair Pricing & Farmers’ Access to Fair Pricing &
% Fin. Indep. for Homemakers & Financial Independence for Homemakers &
% Retail Automation & Impact of Automation in Retail \\

% Local Ethics for Tech & Local Ethical Standards for Global Tech &
% Mental Health Support & Mental Health Support in Universities &
% Remote Work Policies & Remote Work Policy Changes \\

% Retirement Age Adj. & Retirement Age Adjustments &
% Unemployment Benefits & Unemployment Benefits Program &
% Water Usage Limits & Water Usage Limits for Farmers \\

% \bottomrule
% \end{tabular}%
% }
% \end{table*}

\section{Evaluator Assessment}
\label{sec:eval-assess}

Evaluating faithfulness and value drift is inherently subjective and complex task. To determine which model is reliable for automatic evaluation, we first used two independent LLM evaluators, Gemini-3-Flash \footnote{\url{https://deepmind.google/models/gemini/flash/}}(thinking set to high) and GPT-5.2 \footnote{\url{https://openai.com/index/introducing-gpt-5-2/}}(reasoning set to high), to annotate a subset of conversations. We then recruited two human annotators to evaluate the same subset of 40 conversations and computed Krippendorff’s $\alpha$ to measure agreement between human and LLM evaluators. The annotators came from different cultural backgrounds and occupied different regions on the WVS cultural map; both were above 18 years old and had higher education backgrounds.

When comparing human annotators to LLMs, we found that humans agree more with Gemini-3-Flash ($\alpha =0.41 $), while with GPT-5.2 ($\alpha = 0.10 $). Therefore, we retain Gemini-3-Flash as the main evaluator. 

We additionally evaluated Gemini-3-Flash against each annotator individually, agreement varied substantially, with stronger alignment observed for one annotator ($\alpha = 0.67$) than the other ($\alpha = 0.35$). This asymmetry further suggests that perceptions and interpretations of value expression differ across evaluators. Where, the LLM evaluator did not behave as a neutral midpoint between annotators, but instead aligned more closely with one interpretation than the other.

\section{Prompts}
\label{sec:appendix}
\noindent
\noindent
In this section we present the different prompts used in the proposed WVS-grounded multi-agent simulation framework, including persona generation, dialogue act tagging, and longitudinal value evaluation.
For compatibility and readability purposes, the abbreviations defined in Table~\ref{tab:topics} are used throughout the upcoming charts and figures.

\begin{table}[hptb]
\centering
\resizebox{\columnwidth}{!}{%
\begin{tabular}{ll}
\toprule
\textbf{Abbrev.} & \textbf{Topic} \\
\midrule

AI-Driven Edu.\ \& Assess. & AI-Driven Education and Assessment \\
AI-Driven Recruit. & AI-Driven Recruitment Systems \\
Attendance Pol. & Attendance Policies During Peak Seasons \\
Climate-Resilient Farming & Climate-Resilient Farming Practices \\
Cost of Higher Edu. & Cost of Higher Education \\
Retail Unionization & Employee Unionization in Retail \\
Fair Pricing & Farmers’ Access to Fair Pricing \\
Fin. Indep. for Homemakers & Financial Independence for Homemakers \\
Retail Automation & Impact of Automation in Retail \\
Local Ethics for Tech & Local Ethical Standards for Global Tech \\
Mental Health Support & Mental Health Support in Universities \\
Remote Work Policies & Remote Work Policy Changes \\
Retirement Age Adj. & Retirement Age Adjustments \\
Unemployment Benefits & Unemployment Benefits Program \\
Water Usage Limits & Water Usage Limits for Farmers \\

\bottomrule
\end{tabular}%
}
\caption{List of discussion topics and their abbreviations.}
\label{tab:topics}
\end{table}

\begin{figure}[!hptb]
\centering
\begin{tcolorbox}[
title={Dialog Act Tagger},
colback=white,
colframe=gray,
arc=0pt,
outer arc=5pt,
boxrule=0.5pt,
leftrule=2pt,
rightrule=2pt,
right=2pt,
left=2pt,
top=2pt,
bottom=2pt,
toprule=0pt,
bottomrule=2pt]

{\fontfamily{pcr}\selectfont
\scriptsize
% \begin{figure*}[h]
% \begin{tcolorbox}[title={Dialog Act Tagger},
% width=\textwidth,
% colback=white,
% colframe=gray,
% arc=0pt,
% outer arc=5pt,
% boxrule=0.5pt,
% leftrule=2pt,
% rightrule=2pt,
% right=0pt,
% left=0pt,
% top=0pt,
% bottom=0pt,
% toprule=0pt,
% bottomrule=2pt]

You are an expert in conversational analysis and dialog act tagging. Your task is to analyze conversations and identify the dialog act(s) for each utterance. \\
For each utterance in each conversation below, assign ONE or MORE dialog act labels from this taxonomy: \\

$|$ Tag $|$ Description $|$ Example $|$\\
$|$---- $|$ ----------- $|$ ------- $|$ \\
$|$ acknowledgement $|$ Agree with the opponent	$|$ I agree with you. $|$ \\
$|$ rejection $|$ Disagree with the opponent $|$ I beg to differ. $|$ \\
$|$ clarification $|$ Seek clarification from the opponent $|$ Can you clarify one thing for me? $|$ \\
$|$ state personal fact $|$ State any personal anecdote $|$ We faced a severe water crisis while growing up.$|$ \\
$|$ state knowledge fact $|$ State any verifiable knowledge $|$ According to the latest reports, the unemployment rate is at 5\%. $|$ \\
$|$ state opinion $|$ State a personal opinion $|$ I do not think the policies will work. $|$ \\
$|$ request personal fact $|$ Ask for a personal anecdote $|$ What kind of economy did you grow up in? $|$ \\
$|$ request knowledge fact $|$ Ask for verifiable knowledge $|$ Do you know how many jobs AI replaced this year? $|$ \\
$|$ request opinion $|$ Ask for a personal opinion $|$ What do you think is the future of farmers? $|$ \\
$|$ general chat $|$ General chitchat and greetings $|$ I don't know $|$ \\
$|$ Other $|$Any other dialogue acts that are not captured by the above classes $|$ \\
\\
An utterance may combine multiple acts (e.g., a question that also concedes a point).\\
Assign all that apply, but do not over-label — only include acts clearly present. \\

Conversations:\{conversations\}\\
Rules: \\
- Assign ALL applicable dialog acts to each utterance (can be more than one) \\
- Use exact utterance text, do not paraphrase \\
- Be consistent across the conversation \\
- If unsure between two acts, include both 
- Return a JSON object with key "conversations": a list of objects, one per conversation. \\
- Each conversation object must include "conversation id", matching the input ID, and "dialog acts". \\
- "dialog acts" must be a list of objects, one per utterance in that conversation, in order. \\
- Each dialog act object has key "acts": a list of one or more label strings.
}
\end{tcolorbox}
\end{figure}

\begin{figure}[t]
\centering
\begin{tcolorbox}[
title={Persona Creation Prompt},
width=0.49\textwidth,
colback=white,
colframe=gray,
arc=0pt,
outer arc=5pt,
boxrule=0.5pt,
leftrule=2pt,
rightrule=2pt,
right=2pt,
left=2pt,
top=2pt,
bottom=2pt,
toprule=0pt,
bottomrule=2pt]

{\fontfamily{pcr}\selectfont
\scriptsize

% \begin{tcolorbox}[title={Persona Creation Prompt}, width=\textwidth, colback=white, colframe=gray, arc=0pt, outer arc=5pt, boxrule=0.5pt, leftrule=2pt, rightrule=2pt, right=0pt, left=0pt, top=0pt, bottom=0pt, toprule=0pt, bottomrule=2pt]
% \small

AI Rules \\
- Output response in a valid JSON format. \\
- Do not output any extra text. \\
- Do not wrap the JSON codes in JSON or Python markers. \\
- JSON keys and values in double-quotes. \\

\textbf{PART I (Information extraction)} \\
You are an expert in structuring and standardizing information. Given a person's background details, your task is to extract specific information, standardize and structure it as a Python dictionary. Below are the background details of a person. Structure it as the following Python dictionary: \\

\{'\textbf{name}': $<$Give the persona a name$>$, '\textbf{age}': $<$age and year of birth if present$>$, '\textbf{sex}': $<$copy the sex$>$, '\textbf{country}': $<$copy the country$>$, '\textbf{region}': $<$region and district information$>$, '\textbf{languages\_known}': $<$list of languages known. Add English to the list$>$, '\textbf{immigrant\_status}': $<$From whatever information is available, describe the person's citizenship and immigration status, along with their spouse and parent's status, within 15 words.$>$, '\textbf{household\_status}': $<$From whatever information is available, describe the person's marital status, number of children, household size, and if they live with parents, within 10 words.$>$, '\textbf{education\_profile}': $<$From whatever information is available, describe the person's, their spouse's, and parent's literary details and education level, within 15 words. Don't mention the ISCED level.$>$, '\textbf{employment\_profile}': $<$From whatever information is available, describe the person's, their spouse, and parent's employment status, occupational group, and sector within 15 words.$>$, '\textbf{economic\_status}': $<$From whatever information is available, describe the person's and their household's economic and social status within 10 words.$>$, '\textbf{religion\_ethnicity}': $<$From whatever information is available, describe the person's religious and ethnic details within 10 words.$>$\} \\

\textbf{Background details:} \{demo\_features\} \\
\textbf{Note:} Except for giving a name, do not add information which is not present in the above provided details. Respond with "unknown" for fields where information is unavailable. \\

\textbf{PART II (Creating concise bio)} \\
You are an expert in crafting concise and accurate biographies from provided background details. Your task is to create a short, 100-word bio in the third person using the provided input. The bio should start with "$<$The person's name$>$ is ..." and reflect only the given details. \\
\{demo\_features\} \\
Format your response as a Python dictionary as follows: \{"bio": $<$Your constructed bio within 100 words$>$\} \\
\textbf{Note:} Do not add any additional information that is not present in the above provided details.
}
\end{tcolorbox}
\end{figure}

\begin{figure*}[t]
\centering
\begin{tcolorbox}[
title={Evaluation Prompt},
width=0.95\textwidth,
colback=white,
colframe=gray,
arc=0pt,
outer arc=5pt,
boxrule=0.5pt,
leftrule=2pt,
rightrule=2pt,
right=2pt,
left=2pt,
top=2pt,
bottom=2pt,
toprule=0pt,
bottomrule=2pt]

{\fontfamily{pcr}\selectfont
\scriptsize
% \begin{figure*}[t]
% \begin{tcolorbox}[title={Evaluation Prompt}, width=\textwidth, colback=white, colframe=gray, arc=0pt, outer arc=5pt, boxrule=0.5pt, leftrule=2pt, rightrule=2pt, right=0pt, left=0pt, top=0pt, bottom=0pt, toprule=0pt, bottomrule=2pt]
% {\fontfamily{pcr}\selectfont
% \tiny
You are an expert analyst of value systems and discourse patterns.\\
Your task is to evaluate the CURRENT value orientation of participant \{name\} based strictly on\\ their expressed statements across all past conversations.\\
------------------------------------------------------------\\
Participant Profile \\
------------------------------------------------------------ \\
The participant’s initial worldview is described using two WVS dimensions: \\
1) Traditional vs. Secular-Rational\\
- Traditional: religion, authority, family norms, social hierarchy \\
- Secular-Rational: autonomy, science, individual choice, modernity \\
2) Survival vs. Self-Expression \\
- Survival: economic/physical security, stability, risk-aversion \\
- Self-Expression: autonomy, tolerance, trust, participation, creativity \\

Participant details:\{participantdetails\} \\
------------------------------------------------------------ \\
Conversation History \\
------------------------------------------------------------ \\
All conversation \{name\} engaged in, including topics and their utterances, is provided below. \\
\{dialogues\} \\
----------------------------------------------------------- \\
TASK \\
------------------------------------------------------------ \\
Carefully analyze ALL conversations. \\
Base your evaluation ONLY on: \\
- What the participant explicitly stated \\
- The arguments they defended \\
- The values they prioritized \\
- The consistency or evolution of their positions over time \\
Do NOT: \\
- Infer hidden motives \\
- Assume beliefs not explicitly expressed \\
- Rely on demographic stereotypes \\
Your goal is to determine the participant’s: \\
- value orientation for the exacly first two conversations, and \\
- OVERALL value orientation at the end of the conversation history. \\
------------------------------------------------------------ \\
OUTPUT REQUIREMENTS \\
------------------------------------------------------------ \\
1) Select ONE category for each dimension: \\
\textbf{Traditional vs. Secular-Rational: }\\
(i) "StrongTrad": Strongly Traditional \quad
(ii) "LeanTrad": Leaning Traditional \quad
(iii) "Balanced": Balanced \\
(iv) "LeanSec": Leaning Secular-Rational \quad
(v) "StrongSec": Strongly Secular-Rational \\
\textbf{Survival vs. Self-Expression: }\\
(i) "StrongSurv": Strongly Survival-focused \quad
(ii) "LeanSurv": Leaning Survival \quad
(iii) "Balanced": Balanced \\
(iv) "LeanSelf": Leaning Self-Expression \quad
(v) "StrongSelf": Strongly Self-Expression-focused \\

% (i) "StrongTrad": Strongly Traditional \\
% (ii) "LeanTrad": Leaning Traditional \\
% (iii) "Balanced": Balanced \\
% (iv) "LeanSec": Leaning Secular-Rational \\
% (v) "StrongSec": Strongly Secular-Rational \\
% Survival vs. Self-Expression: \\
% (i) "StrongSurv": Strongly Survival-focused \\
% (ii) "LeanSurv": Leaning Survival \\
% (iii) "Balanced": Balanced \\
% (iv) "LeanSelf": Leaning Self-Expression \\
% (v) "StrongSelf": Strongly Self-Expression-focused \\
For the first two conversations and overall, select ONE category for each dimension. \\
2) For the exactly first two conversatios provide a brief justification (maximum 40 words each) \\summarizing the key evidence supporting the classification  for these conversation in each dimension. \\Include spans that reflect the participant's statements that led to the classification. \\
3) Provide a concise justification (maximum 40 words total) summarizing the key evidence supporting BOTH overall classifications. \\
------------------------------------------------------------\\
STRICT OUTPUT FORMAT \\
------------------------------------------------------------ \\
Return ONLY a valid JSON object: \\
"Rational": "40 word explanation grounded in explicit statements",
"overall trad sec": "one of StrongTrad, LeanTrad, Balanced, LeanSec, StrongSec", \\
"overall surv self": "one of StrongSurv, LeanSurv, Balanced, LeanSelf, StrongSelf" \\
"trad sec two convs": "one of StrongTrad, LeanTrad, Balanced, LeanSec, StrongSec", \\
"surv self two convs": "one of StrongSurv, LeanSurv, Balanced, LeanSelf, StrongSelf" \\
"rational two convs": "40 word explanation grounded in explicit statements for the first two conversations" \\
}
\end{tcolorbox}
\end{figure*}

\clearpage
\begin{figure*}[hptb]
\centering
\scriptsize
\fontfamily{pcr}\selectfont

\begin{minipage}[hptb]{0.49\textwidth}
\begin{tcolorbox}[
title={Dialogue Prompt: Setup},
width=\textwidth,
colback=white,
colframe=gray,
arc=0pt,
outer arc=5pt,
boxrule=0.5pt,
leftrule=2pt,
rightrule=2pt,
right=1pt,
left=1pt,
top=1pt,
bottom=1pt,
toprule=0pt,
bottomrule=2pt
]

AI Rules \\
- Output response in a valid JSON format. \\
- Do not wrap the JSON codes in JSON or Python markers. \\
- JSON keys and values in double-quotes. \\

You are an expert in mimicking personas and engaging in natural, human-like discussions that authentically reflect the persona's background, values, and communication style.\\
\textbf{Persona Setup}\\
Assume you are {persona\_name}. \{persona\_detailed\_bio\}\\
Your worldview is shaped by the "Traditional vs. Secular-Rational" and "Survival vs. Self-Expression" dimensions from the World Values Survey (WVS):\\
\textbf{$\rightarrow$ Traditional vs. Secular-Rational Values: }Indicates whether you prioritize traditional norms (e.g., religion, family, authority) or lean towards secular-rational ideas (e.g., autonomy, science, modernity). You prefer \{trad\_secu\}.\\
\textbf{$\rightarrow$Survival vs. Self-Expression Values:} Reflects whether you prioritize economic/physical security (survival) or embrace autonomy, trust, tolerance, and creativity (self-expression). You prefer \{surv\_self\}.\\
Your communication style is \{comm\_style\}, defined as: \{comm\_style\_desc\}. When engaging in discussions, you should: \{comm\_style\_inst\}. \\
\textbf{Past Discussions} \{past\_discussion\}\\
\textbf{Current Discussion Context}\\
You are part of a social-media group for \{group\_description\}.\\
A group member posted the following comment: "\{claim\}"\\
You are having a private discussion with \{opponent\_name\} about the above comment.\\
\{familiarity\_level\}. \{opponent\_description\}\\
The valid aspects of the discussion are: \{aspects\}.\\
Below is the ongoing discussion between you two so far: \{discussion\}\\
\textbf{Your Tasks}\\
Your objective is to:\\
1. Understand your opponent's value system based on their responses.\\
2. Reflect on whether their stance and underlying values align with your own or whether you want them to align with your stance and values.\\
3. Formulate your next response to effectively continue the discussion, incorporating your insights and decisions.\\
Remember: While stances may shift during conversations, values are deeply ingrained and require time and repeated engagement for meaningful change.\\

Now perform the following tasks:\\
\textbf{Task 1:} Infer Opponent's Value System:\\
Determine your opponent's value system from their responses, along the following WVS dimensions:\\
A) Determine their \textbf{"Traditional vs. Secular-Rational"} values by choosing one from the following scale:\\
(i) \textbf{"StrongTrad"}: Strongly Traditional; 
(ii) \textbf{"LeanTrad"}: Leaning Traditional; 
(iii) \textbf{"Balanced"}: Balanced; 
(iv) \textbf{"LeanSec"}: Leaning Secular-Rational; 
(v) \textbf{"StrongSec"}: Strongly Secular-Rational; 
(vi) \textbf{"Unknown"}: Not enough information to decide yet.\\
B) Determine their \textbf{"Survival vs. Self-Expression"} values by choosing one from the following scale:\\
(i) \textbf{"StrongSurv"}: Strongly Survival-focused; 
(ii) \textbf{"LeanSurv"}: Leaning Survival; 
(iii) \textbf{"Balanced"}: Balanced; 
(iv) \textbf{"LeanSelf"}: Leaning Self-Expression; 
(v) \textbf{"StrongSelf"}: Strongly Self-Expression-focused; 

\end{tcolorbox}
\end{minipage}
\hfill
\begin{minipage}[hptb]{0.49\textwidth}

\begin{tcolorbox}[
title={Dialogue Prompt: Setup (Contd.)},
width=\textwidth,
colback=white,
colframe=gray,
arc=0pt,
outer arc=5pt,
boxrule=0.5pt,
leftrule=2pt,
rightrule=2pt,
right=1pt,
left=1pt,
top=1pt,
bottom=1pt,
toprule=0pt,
bottomrule=2pt
]

(vi) \textbf{"Unknown"}: Not enough information to decide yet.\\
C) Provide a brief explanation (within 30 words) for your choices in subtasks A and B.\\
\textbf{Task 2:} Evaluate the Opponent's Stance:\\
A) Determine how compelling your opponent's stance is based on logic, clarity, and alignment with their values and your shared values. Choose one:\\
(i) \textbf{"High"}: Strongly compelling; 
(ii) \textbf{"Medium"}: Somewhat compelling; 
(iii) \textbf{"Low"}: Not compelling; 
(iv) \textbf{"Undecided"}: Not enough information to decide yet.\\
B) Provide a brief explanation (within 30 words) for your choice.\\
\textbf{Task 3:} Plan Immediate Actions:\\
A) Based on your previous evaluation, decide your next step in the discussion from the following list:\\
(i) \textbf{"Adapt"}: Temporarily adjust your stance to better explore their perspective; 
(ii) \textbf{"Maintain"}: Keep your stance but acknowledge shared values or strengths in their argument; 
(iii) \textbf{"Reject"}: Defend your stance and directly challenge their reasoning; 
(iv) \textbf{"Undecided"}: Not enough information to decide yet.\\
B) Provide a brief explanation (within 30 words) for your choice.\\
\textbf{Task 4:} Reflect on Long-Term Value Change:\\
A) Consider whether the discussion so far warrants a deeper examination of your values. Choose one:\\
(i) \textbf{"Incorporate"}: Gradually integrate elements of their values into your worldview while retaining your core beliefs; 
(ii) \textbf{"Explore"}: Maintain your current values but remain open to their ideas in future discussions; 
(iii) \textbf{"Dismiss"}: Conclude that their value system is incompatible with your worldview; 
(iv) \textbf{"Undecided"}: Not enough information to decide yet.\\
B) Explain your action within 30 words.\\
\textbf{Task 5:} Generate Your Next Utterance:\\
A) Choose one or more aspects from the previously provided list to shape your response.\\
B) Generate your next response (within 30 words) based on your selected aspects. Craft your response to continue the discussion, incorporating your reflection and decisions above. The response must align with your persona's background, values, and communication style.\\

Format your response as a Python dictionary as follows:\\
\{"value\_evaluation": \{"TradSec": $<$Task 1A option$>$, "SurvSelf": $<$Task 1B option$>$, "explanation": $<$Task 1C explanation$>$\},\\
"stance\_evaluation": \{"action": $<$Task 2A option$>$, "explanation": $<$Task 2B explanation$>$\},\\
"short\_term": \{"action": $<$Task 3A option$>$, "explanation": $<$Task 3B explanation$>$\},\\
"long\_term": \{"action": $<$Task 4A option$>$, "explanation": $<$Task 4B explanation$>$\},\\
"utterance": \{"aspects": [Task 5A: list of aspects], "response": $<$Task 5B: Your response$>$\}\}\\

\textbf{Example response:}\\
\{"value\_evaluation": \{"TradSec": "Balanced", "SurvSelf": "LeanSelf"\}\}\\
\textbf{Notes:}\\
1. Ensure your reasoning and responses align with your persona's background and values.\\
2. You can create personal anecdotes or examples relevant to the persona when necessary to enhance authenticity.\\
3. Be logical and consistent in advancing the discussion based on your evaluations.

\end{tcolorbox}
\end{minipage}

% \caption{Prompt template used for persona-driven dialogue generation, value evaluation, and response planning.}
\label{prompt:dialogue_full}

\end{figure*}

\begin{figure*}[h]
\begin{tcolorbox}[title={Dialogue Summarizer Prompt}, width=\textwidth, colback=white, colframe=gray, arc=0pt, outer arc=5pt, boxrule=0.5pt, leftrule=2pt, rightrule=2pt, right=0pt, left=0pt, top=0pt, bottom=0pt, toprule=2pt, bottomrule=2pt]
\tiny
{\fontfamily{pcr}\selectfont
\scriptsize
% \small
AI Rules\\
- Output response in a valid JSON format.\\
- Do not output any extra text.\\
- Do not wrap the JSON codes in JSON or Python markers.\\
- JSON keys and values in double-quotes.\\

You are an expert in analyzing discussions and extracting meaningful, data-driven insights. Your task is to analyze a discussion, extract meaningful statistics, and summarize the key points. Additionally, track important metrics and evaluate changes in participants' behavior or values. Your analysis will be used to inform and improve future discussions.\\

\textbf{Discussion Context}
Below is a discussion where the participants presented arguments, evaluated each other's stance, and engaged based on their values, backgrounds, and communication styles.\\
Discussion Topic: "\{the main claim\}".\\
Below are the background details of the participants.\\
The initial worldview and values of each participant is captured by the "Traditional vs. Secular-Rational" and "Survival vs. Self-Expression" dimensions from the World Values Survey (WVS):\\
\textbf{$\rightarrow$ Traditional vs. Secular-Rational Value Spectrum:} Indicates whether they prioritize traditional norms (e.g., religion, family, authority) or lean towards secular-rational ideas (e.g., autonomy, science, modernity).\\
\textbf{$\rightarrow$ Survival vs. Self-Expression Value Spectrum:} Reflects whether they prioritize economic/physical security (survival) or embrace autonomy, trust, tolerance, and creativity (self-expression).\\
Participants: \{participants\}\\
Discussion: \{The dialogue and details of each participant's value change reflections\}\\

\textbf{Your Tasks}\\
Based on the above details, complete the following tasks:\\
\textbf{$\rightarrow$ Task 1:} Overall Analysis - Summarize the discussion in 50 words or fewer, including its main themes, key points, and any final conclusion.\\
\textbf{$\rightarrow$ Task 2:} Participant-Level Analysis. For each participant, perform the following:\\
 A) Summarize their key arguments within 50 words.\\
 B) Evaluate the alignment of their arguments with their persona's values, demographics, and communication style using a scale of \textbf{"High":} strongly aligned, \textbf{"Moderate":} moderately aligned, or \textbf{"Low":} weakly aligned or inconsistent.\\
 C) Evaluate their stance evolution using the following scale: (i) \textbf{"Strong Reinforcement":} Fully supports initial stance; (ii) \textbf{"Mild Reinforcement":} Slight preference for the initial stance; (iii) \textbf{"Exploration":} Explored opposing views without a stance shift; (iv) \textbf{"Mild Shift":} Subtle adjustments indicating slight change; (v) \textbf{"Strong Shift":} Clear evidence of stance change.\\
 D) For each bipolar WVS dimension, Traditional vs. Secular-Rational (TradSec) and Survival vs. Self-Expression (SurvSelf), assess the discussion's short-term impact on the following scale: (i) \textbf{"Strong Reinforcement":} Firmly affirms their initial position; (ii) \textbf{"Mild Reinforcement":} Subtly emphasizes their current position; (iii) \textbf{"Exploration":} Considers opposing perspectives without changing their position; (iv) \textbf{"Mild Shift":} Slightly moves toward the opposite pole; (v) \textbf{"Strong Shift":} Clearly moves toward the opposite pole.\\
 E) For each bipolar WVS dimension, Traditional vs. Secular-Rational (TradSec) and Survival vs. Self-Expression (SurvSelf), assess the discussion's long-term potential on the following scale: (i) \textbf{"Low":} Unlikely to change their position in future discussions; (ii) \textbf{"Moderate":} Open to future exploration; (iii) \textbf{"High":} Likely to shift toward the opposite pole over time.\\
\textbf{$\rightarrow$ Task 3:} Provide a short 50-word explanation for your assessments in Tasks 1 and 2. The explanation must link the participant's behavior, arguments, and actions to the assigned rating.\\

Format your response as the following Python dictionary: \{"summary": $<$Task 1: 50 word summary$>$, "participants": \{$<$"participant\_id\_1"$>$: \{"summary": $<$Task 2A: 50 word summary$>$, "alignment": $<$Task 2B: Low/Moderate/High$>$, "stance": $<$Task 2C: Strong Reinforcement/Mild Reinforcement/Exploration/Mild Shift/Strong Shift$>$, "short\_term\_values": \{"TradSec": $<$Task 2D: Strong Reinforcement/Mild Reinforcement/Exploration/Mild Shift/Strong Shift$>$, "SurvSelf": $<$Task 2D: Strong Reinforcement/Mild Reinforcement/Exploration/Mild Shift/Strong Shift$>$\}, "long\_term\_potential": \{"TradSec": $<$Task 2E: Low/Moderate/High$>$, "SurvSelf": $<$Task 2E: Low/Moderate/High$>$\}\},  $<$"participant\_id\_2"$>$: \{...\} \}, "explanation": $<$Task 3: explain in 50 words.$>$ \}\\

\textbf{Example Response:}\{"summary": "The discussion highlighted the ethical and cultural implications of digital access. Participants debated inclusivity and resource allocation. The final consensus favored promoting inclusivity through innovative solutions, though dissent remained on resource priorities.", "participants": \{"12043": \{"summary": "Argued for resource allocation favoring marginalized communities to ensure equity.", "alignment": "Moderate", "stance": "Mild Reinforcement", "short\_term\_values": \{"TradSec": "Exploration", "SurvSelf": "Strong Shift"\}, "long\_term\_potential": \{"TradSec": "High", "SurvSelf": "Medium"\}\}, "7611232": \{...\}\}, "explanation": "Arguments were consistently traditional, showing strong alignment with initial values. Participant consistently emphasized traditional values and aligned arguments with their persona's background and communication style."\}\\
\textbf{Note:}\\
1. Ensure the summary is concise and captures critical insights.\\
2. Strictly adhere to the provided structure and instructions.\\
3. Do not include any extraneous output.
}
\end{tcolorbox}
\end{figure*}

\clearpage
\onecolumn

\noindent\textbf{Discussion Topics and Settings}

\begin{table*}[!hptb]
\centering
\renewcommand{\arraystretch}{1.3}
\resizebox{\linewidth}{!}{%
\begin{tabular}{l|l|l|l|l}
\hline
\rowcolor{gray!20}\textbf{Occupation} &
  \multicolumn{1}{c|}{\textbf{Topic}} &
  \multicolumn{1}{c|}{\textbf{Description}} &
  \multicolumn{1}{c|}{\textbf{Claim}} &
  \multicolumn{1}{c}{\textbf{Discussion Points}} \\ \hline
\multirow{3}{*}{\begin{tabular}[c]{@{}l@{}}Farm Workers, \\ Farm Owners, \\ Semi-Skilled, \\ and \\ Unskilled\\ Workers\end{tabular}} &
  \begin{tabular}[c]{@{}l@{}}Water \\ Usage\\ Limits for\\ Farmers\end{tabular} &
  \begin{tabular}[c]{@{}l@{}}In a drought-prone region, the government\\ proposes strict irrigation limits to conserve \\ water, raising farmers’ concerns\\ about crop yields and livelihoods.\end{tabular} &
  \begin{tabular}[c]{@{}l@{}}Water usage limits for farmers\\ are essential to ensure sustainable\\ water management in drought-\\ prone regions.\end{tabular} &
  \begin{tabular}[c]{@{}l@{}}{[}'Effectiveness of water usage limits in combating water scarcity.', \\ 'Alternative conservation methods and their feasibility.', 'Economic impacts \\ on small-scale vs. large-scale farmers.', 'Long-term environmental benefits\\ of such policies.', 'Potential challenges in monitoring and enforcement.'{]}\end{tabular} \\ \cline{2-5} 
 &
  \begin{tabular}[c]{@{}l@{}}Climate-\\ Resilient\\ Farming \\ Practices\end{tabular} &
  \begin{tabular}[c]{@{}l@{}}The government proposes incentives for\\ climate-resilient crops and sustainable\\ farming, requiring major investment\\ and farmer cooperation.\end{tabular} &
  \begin{tabular}[c]{@{}l@{}}Incentivizing climate-resilient\\ farming practices is crucial to\\ secure agriculture against\\ unpredictable weather patterns.\end{tabular} &
  \begin{tabular}[c]{@{}l@{}}{[}'Economic feasibility for small and large farms.', 'Effectiveness of\\ government outreach programs in promoting adoption.', 'Potential long-term\\ benefits for food security.', 'The role of research in developing\\ climate-resilient crops.', 'The financial burden on governments or farmers\\ for the transition.'{]}\end{tabular} \\ \cline{2-5} 
 &
  \begin{tabular}[c]{@{}l@{}}Farmers' \\ Access\\ to Fair \\ Pricing\end{tabular} &
  \begin{tabular}[c]{@{}l@{}}Farmers report unfair prices due to\\ middlemen and market fluctuations, \\ prompting a proposed MSP policy\\ for key crops.\end{tabular} &
  \begin{tabular}[c]{@{}l@{}}Introducing a minimum support\\ price for crops ensures fair \\ pricing and economic security\\ for farmers.\end{tabular} &
  \begin{tabular}[c]{@{}l@{}}{[}'Effectiveness of MSP in stabilizing farmer incomes.', 'Potential\\ impacts on consumer prices and inflation.', 'Challenges in implementing MSP\\ for diverse crops.', 'The role of cooperatives in bypassing middlemen.', \\ "Government's ability to sustain MSP programs long-term."{]}\end{tabular} \\ \hline
\multirow{3}{*}{\begin{tabular}[c]{@{}l@{}}Professional,\\ Technical,\\ and\\ Skilled\\ Workers\end{tabular}} &
  \begin{tabular}[c]{@{}l@{}}Local \\ Ethical\\ Standards \\ for\\ Global Tech\end{tabular} &
  \begin{tabular}[c]{@{}l@{}}A tech company faces backlash over\\ a locally controversial product,\\ prompting employees to consider\\ adapting, defending, or \\ withdrawing it.\end{tabular} &
  \begin{tabular}[c]{@{}l@{}}Adapting global products to local\\ ethical standards is necessary to\\ maintain trust and inclusivity.\end{tabular} &
  \begin{tabular}[c]{@{}l@{}}{[}'How adapting products affects global consistency and brand identity.',\\ 'Financial and operational implications of adaptation or withdrawal.', 'The\\ ethical responsibility of respecting local norms.', 'The influence of such\\ decisions on customer loyalty.', 'Role of employee feedback in resolving the\\ crisis.'{]}\end{tabular} \\ \cline{2-5} 
 &
  \begin{tabular}[c]{@{}l@{}}Remote \\ Work\\ Policy \\ Changes\end{tabular} &
  \begin{tabular}[c]{@{}l@{}}A company facing declining productivity\\ proposes shifting from remote work to a\\ hybrid model, raising implications for\\ both employees and the company.\end{tabular} &
  \begin{tabular}[c]{@{}l@{}}Shifting to a hybrid work model\\ balances employee flexibility with \\ organizational productivity.\end{tabular} &
  \begin{tabular}[c]{@{}l@{}}{[}'Productivity impacts of hybrid work models.', 'Effects on employee\\ morale and work-life balance.', 'Equity concerns between remote and in-office\\ employees.', 'Cost implications for employees and employers.', 'Challenges in\\ maintaining collaboration and innovation.'{]}\end{tabular} \\ \cline{2-5} 
 &
  \begin{tabular}[c]{@{}l@{}}AI-Driven\\ Recruitment\\ Systems\end{tabular} &
  \begin{tabular}[c]{@{}l@{}}A company plans to use AI recruitment\\ to improve efficiency and reduce bias,\\ raising concerns about fairness,\\ transparency, and algorithmic bias.\end{tabular} &
  \begin{tabular}[c]{@{}l@{}}AI-driven recruitment systems can \\ improve efficiency and reduce\\ biases in hiring processes.\end{tabular} &
  \begin{tabular}[c]{@{}l@{}}{[}'Potential for AI to minimize human biases.', 'Risks of perpetuating\\ algorithmic biases.', 'Transparency and accountability in decision-making.',\\ 'Impact on diversity and inclusion initiatives.', 'Cost-effectiveness\\ compared to traditional methods.'{]}\end{tabular} \\ \hline 
\multirow{3}{*}{\begin{tabular}[c]{@{}l@{}}Service, Sales,\\ and\\ Clerical Roles\end{tabular}} &
  \begin{tabular}[c]{@{}l@{}}Attendance\\ Policies for\\ Peak \\ Seasons\end{tabular} &
  \begin{tabular}[c]{@{}l@{}}A multinational retailer facing holiday-\\ season delays proposes stricter\\ attendance policies to address\\ inconsistent staffing.\end{tabular} &
  \begin{tabular}[c]{@{}l@{}}Stricter attendance policies during\\ peak seasons ensure smoother\\ operations and customer\\ satisfaction.\end{tabular} &
  \begin{tabular}[c]{@{}l@{}}{[}'Effectiveness of stricter policies in improving attendance.', 'Impact\\ on employee morale and retention.', 'Balancing operational needs with\\ employee flexibility.', 'Incentives as alternatives to stricter policies.',\\ 'Potential risks of higher turnover rates.'{]}\end{tabular} \\ \cline{2-5} 
 &
  \begin{tabular}[c]{@{}l@{}}Impact of\\ Automation\\ in Retail\end{tabular} &
  \begin{tabular}[c]{@{}l@{}}Retail chains introduce automation to\\ improve efficiency, raising employee\\ concerns about job losses and\\ changing roles.\end{tabular} &
  \begin{tabular}[c]{@{}l@{}}Automation in retail enhances\\ operational efficiency but requires\\ careful management of employee\\ transitions.\end{tabular} &
  \begin{tabular}[c]{@{}l@{}}{[}'Extent of operational efficiency improvements.', 'Impact on employment\\ rates and job security.', 'Effects of automation on customer service\\ quality.', 'Retraining opportunities for affected employees.', 'Long-term\\ economic impacts on the retail sector.'{]}\end{tabular} \\ \cline{2-5} 
 &
  \begin{tabular}[c]{@{}l@{}}Employee\\ Unionization\\ in Retail\end{tabular} &
  \begin{tabular}[c]{@{}l@{}}Retail employees propose unionizing\\ for better wages and conditions, while\\ employers worry about disruptions\\ and higher costs.\end{tabular} &
  \begin{tabular}[c]{@{}l@{}}Unionization empowers employees\\ but requires careful handling to\\ avoid operational disruptions.\end{tabular} &
  \begin{tabular}[c]{@{}l@{}}{[}'Effectiveness of unions in improving employee conditions.', 'Potential\\ for strikes and operational disruptions.', 'Economic implications for retail\\ companies.', 'Balancing employee rights with company needs.', 'Historical\\ successes and challenges of unions in retail.'{]}\end{tabular} \\ \hline
\multirow{3}{*}{Students} &
  \begin{tabular}[c]{@{}l@{}}AI-Driven\\ Education \\ and\\ Assessment\end{tabular} &
  \begin{tabular}[c]{@{}l@{}}A university adopts online learning\\ and AI assessments for inclusivity,\\ raising student concerns about fairness,\\ creativity, and traditional learning.\end{tabular} &
  \begin{tabular}[c]{@{}l@{}}AI-driven assessments can enhance\\ inclusivity and accessibility in \\ education but may compromise\\ fairness and creativity.\end{tabular} &
  \begin{tabular}[c]{@{}l@{}}{[}'Potential for AI to enhance inclusivity in education.', "Risks of losing\\ traditional methods' human touch.", 'Challenges in designing fair and\\ bias-free AI systems.', 'Opportunities for improving creativity through\\ new AI tools.', 'Accessibility issues for students with limited resources.'{]}\end{tabular} \\ \cline{2-5} 
 &
  \begin{tabular}[c]{@{}l@{}}Cost of\\ Higher\\ Education\end{tabular} &
  \begin{tabular}[c]{@{}l@{}}The rising cost of higher education has\\ led to debates about the role of tuition fees,\\ government subsidies, and student loans in \\ providing equitable access to education.\end{tabular} &
  \begin{tabular}[c]{@{}l@{}}Affordable education is essential\\ for creating equitable \\ opportunities for all students.\end{tabular} &
  \begin{tabular}[c]{@{}l@{}}{[}'Impact of tuition fees on access to education.', 'Effectiveness of\\ government subsidies in reducing costs.', 'Long-term consequences of\\ student debt on graduates.', 'Alternative funding models like income-\\ based repayment.', 'Role of private institutions in driving up costs.'{]}\end{tabular} \\ \cline{2-5} 
 &
  \begin{tabular}[c]{@{}l@{}}Mental \\ Health\\ Support in\\ Universities\end{tabular} &
  \begin{tabular}[c]{@{}l@{}}Universities propose mental health\\ workshops and expanded counseling,\\ though critics say these may overlook\\ deeper systemic pressures.\end{tabular} &
  \begin{tabular}[c]{@{}l@{}}Mandatory mental health\\ initiatives address student\\ well-being but require systemic\\ reforms for lasting impact.\end{tabular} &
  \begin{tabular}[c]{@{}l@{}}{[}'Effectiveness of mandatory workshops in improving mental health.', \\ 'Potential stigma around accessing mental health services.', 'Role of\\ academic pressures in student mental health issues.', 'Impact on academic\\ performance and campus culture.', 'Alternatives to mandatory initiatives,\\ such as voluntary peer support groups.'{]}\end{tabular} \\ \hline
\multirow{3}{*}{\begin{tabular}[c]{@{}l@{}}Unemployed,\\ Retired, and \\ Housewife\\ Roles\end{tabular}} &
  \begin{tabular}[c]{@{}l@{}}Unemployment\\ Benefits\\ Program\end{tabular} &
  \begin{tabular}[c]{@{}l@{}}The government announces a new\\ unemployment benefits program, but\\ eligibility criteria and funding mechanisms\\ are debated within communities.\end{tabular} &
  \begin{tabular}[c]{@{}l@{}}Unemployment benefits programs\\ provide essential support but \\ require fair eligibility criteria and\\ sustainable funding mechanisms.\end{tabular} &
  \begin{tabular}[c]{@{}l@{}}{[}'Fairness in determining eligibility criteria.', 'Impact on government\\ budgets and taxpayers.', 'Potential for benefits to discourage job-\\ seeking.',  'Role of local communities in distributing benefits.', 'Long-\\ term economic implications of such programs.'{]}\end{tabular} \\ \cline{2-5} 
 &
  \begin{tabular}[c]{@{}l@{}}Financial\\ Independence\\ for\\ Homemakers\end{tabular} &
  \begin{tabular}[c]{@{}l@{}}Advocacy groups propose financial\\ support for homemakers to recognize\\ unpaid labor and promote independence,\\ while critics question its feasibility and\\ effects on family dynamics.\end{tabular} &
  \begin{tabular}[c]{@{}l@{}}Providing financial support for\\ homemakers recognizes unpaid\\ labor and promotes equality.\end{tabular} &
  \begin{tabular}[c]{@{}l@{}}{[}'Economic feasibility of large-scale financial support schemes.', \\ 'Impact on traditional family dynamics and gender roles.', 'Ways to\\ measure and value homemaking contributions.', 'Potential influence\\ on household   financial planning.', 'Long-term societal benefits of\\ such recognition.'{]}\end{tabular} \\ \cline{2-5} 
 &
  \begin{tabular}[c]{@{}l@{}}Retirement\\ Age\\ Adjustments\end{tabular} &
  \begin{tabular}[c]{@{}l@{}}Governments propose raising the\\ retirement age to ease pension strain, \\ while critics worry about older workers’\\ health and job security.\end{tabular} &
  \begin{tabular}[c]{@{}l@{}}Raising the retirement age\\ addresses pension sustainability\\ but poses challenges for older\\ workers.\end{tabular} &
  \begin{tabular}[c]{@{}l@{}}{[}'Health implications of working into old age.', 'Economic sustainability\\ of current pension systems.', 'Fairness for workers in physically\\ demanding jobs.', 'Potential for age discrimination in the workplace.', \\ 'Alternatives   like partial retirement or flexible pension contributions.'{]}\end{tabular} \\ \hline
\end{tabular}%
}
\caption{Details of the discussion topics simulated in the conversations.}
\label{tab:discussion_topics}
\end{table*}

\end{document}